%% file: tpm2024.tex
\documentclass[accepted]{tpm2024} 

\usepackage[american]{babel}

\usepackage{natbib}
\usepackage{xcolor}

\usepackage[font=small]{caption}

\usepackage{bm}
\usepackage{amsthm}
\usepackage{amssymb}
\usepackage{amsmath}
\usepackage{amsfonts}
\usepackage{mathtools}
\usepackage{dsfont}
\usepackage{graphicx}

\newtheorem{assumption}{Assumption}
\newtheorem{proposition}{Proposition}

\usepackage{fancyvrb}

\definecolor{c1}{RGB}{27,158,119}
\definecolor{c2}{RGB}{117,112,179}
\definecolor{c3}{RGB}{217,95,2}
\definecolor{c4}{RGB}{231,41,138}
\definecolor{c5}{RGB}{130,130,130}

\usepackage{tikz}
\usetikzlibrary{calc}

\usepackage{balance}
\usepackage{booktabs}

\title{GraphSPNs: Sum-Product Networks Benefit From Canonical Orderings}

\author[1]{\href{mailto:<papezmil@fel.cvut.cz>?Subject=GraphSPNs: Sum-Product Networks Benefit From Canonical Orderings}{Milan Pape\v{z}}{}}
\author[1]{Martin Rektoris}
\author[1]{V\'{a}clav \v{S}m\'{i}dl}
\author[1]{Tom\'{a}\v{s} Pevn\'{y}}
\affil[1]{%
    Artificial Intelligence Center\\
    Czech Technical University\\
    Prague, Czech Republic
}

\begin{document}
\maketitle

\begin{abstract}
Deep generative models have recently made remarkable progress in capturing complex probability distributions over graphs. However, they are intractable and thus unable to answer even the most basic probabilistic inference queries without resorting to approximations. Therefore, we propose graph sum-product networks (GraphSPNs), a tractable deep generative model that provides exact and efficient inference over (arbitrary parts of) graphs. We investigate different principles to make SPNs permutation invariant. We demonstrate that GraphSPNs can (conditionally) generate novel and chemically valid molecular graphs, being competitive to, and sometimes even better than, existing intractable models. We find out that (Graph)SPNs benefit from ensuring the permutation invariance via canonical ordering. 
\end{abstract}

\section{Introduction}
Graphs are a fundamental framework for representing real or abstract objects and their hierarchical interactions in a diverse range of scientific and engineering applications, such as discovering new materials \citep{choudhary2022recent}, developing personalized diagnostic strategies \citep{chandak2023building}, and estimating time of arrival \citep{derrow2021eta}. Nonetheless, designing probabilistic models for graphs is challenging. Graphs can exhibit highly complex and combinatorial structures, making it difficult to capture their probabilistic behavior effectively. While traditional approaches struggle to handle this problem, deep generative models---which rely on expressive graph neural networks \citep{wu2020comprehensive,zhang2020deep}---have recently made significant progress in this direction \citep{you2018graphrnn,simonovsky2018graphvae,de2018molgan,shi2020graphaf,jo2022score}.

The complication with these models is their limited ability to carry out probabilistic inference tasks beyond mere sampling and, typically, exact likelihood evaluation. For example, arbitrary conditioning with an already trained model allows us to sample conditionally on an existing (part of a) graph. The conditioning significantly reduces the space of possible solutions, which is enormous in some domains \citep{reymond2012enumeration}. Other inference mechanisms, including marginalization, maximum a posteriori estimation, and expectation, have great potential in designing graphs with desired profiles.

Sum-product networks (SPNs) \citep{poon2011sum} are deep generative models for \emph{fixed-size} tensor data. Their essential feature is that they are tractable, which means that---under certain assumptions---they guarantee to answer a large family of complex probabilistic queries exactly and efficiently \citep{vergari2021compositional,choi2020probabilistic}. The existing work on SPNs for graphs is limited to representation learning \citep{zheng2018learning,errica2023tractable}. Therefore, we propose GraphSPNs, deep (one-shot) generative models for tractable probabilistic inference on graphs. There are two main challenges in designing such models. (i) SPNs are probability distributions defined on a \emph{fixed-dimensional} space. Hence, we have to deal with the fact that each instance of a graph has a different number of nodes and edges. (ii) Graphs are permutation invariant objects \citep{veitch2015class}. Consequently, our models must be agnostic to re-ordering the nodes in a graph. In other words, the probability of a graph has to remain unchanged when permuting the nodes (we want to recognize the same graph up to all its permutations). We address (i) in a pragmatic way, using virtual nodes. This principle is frequently adopted in deep generative models for graphs \citep{madhawa2019graphnvp}. However, the real difficulty is to deal with (ii). Indeed, learning a permutation invariant distribution is hard since the number of modes of the target data distribution is much higher than a non-invariant, canonical distribution. Therefore, this paper investigates different techniques to make GraphSPNs permutation invariant.

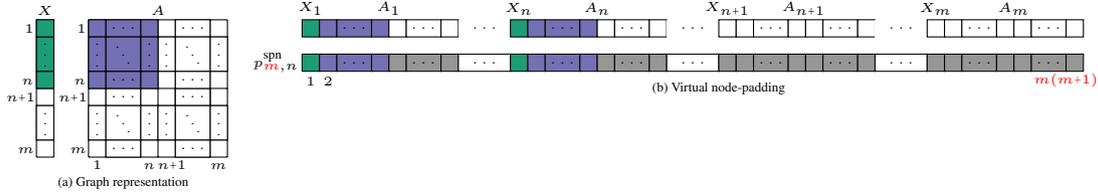
\begin{figure*}
    \centering
    \input{padding.tikz}
    \vspace{-12pt}
    \caption{\emph{Graph representation.} (a) Let $G$ be a graph represented by a feature matrix, $X\in\mathcal{X}^n$, and an adjacency tensor, $A\in\mathcal{A}^{n\times n}$. We consider each instance of $G$ (highlighted in \textcolor{c1}{green} and \textcolor{c2}{blue}) to have a random number of nodes, $n\in(0,1,\ldots,m)$, but we expect it to have at most $m$ nodes. The remaining places (white) are empty and are not included in the training data. (b) Virtual node-padding fills in the empty places with virtual nodes (\textcolor{c5}{grey}), which requires us to extend $\mathcal{X}$ by an extra category, $\mathcal{X}\coloneqq(0,1)^{q+1}$.
    }
    \label{fig:padding}
\end{figure*}

\section{Background}\label{sec:background}

\paragraph{Sum-product networks.} A tensorized SPN \citep{peharz2020einsum,peharz2020random,loconte2024subtractive} is a deep learning model of a probability distribution, $p(X)$, over a \emph{fixed-size} tensor, $X\in\mathcal{X}$. The network contains several layers of computational units (similar to neural networks \citep{vergari2019visualizing}). Each layer is defined over its \emph{scope}, $\psi\subseteq X$, i.e., a subset of the input. There are three types of layers, depending on the units they encapsulate: sum layer $\mathsf{L}_\mathsf{S}$, product layer $\mathsf{L}_\mathsf{P}$, and input layer $\mathsf{L}_\mathsf{I}$. The units of \emph{input} layers are user-defined probability distributions, $p_i(\psi)$. For $n_I$ units, an input layer computes $p_i(\psi)$ for $i\in(1,\ldots,n_I)$ and outputs an $n_I$-dimensional vector of probabilities $\mathbf{l}$. The units of \emph{product} layers are factored distributions, applying conditional independence over a pair-wise disjoint partition of their scope. A product layer receives outputs from $n$ layers, $(\mathbf{l}_1,\ldots,\mathbf{l}_n)$, and computes either an Hadamard product, $\mathbf{l}=\odot^n_{i=1}\mathbf{l}_i$, or Kronecker product, $\mathbf{l}=\otimes^n_{i=1}\mathbf{l}_i$. The units of \emph{sum} layers are mixture distributions. For $n_S$ units, a sum layer receives an $n$-dimensional input, $\mathbf{l}$, from a previous layer and computes $\mathbf{W}\mathbf{l}$, where $\mathbf{W}$ is an $n_S\times n$ matrix of row-normalized weights. The output (layer) of a tensorized SPN is typically a sum layer.

The key benefit of SPNs is that they provide \emph{tractable} inference over arbitrary subsets of $X$. However, to this end, the sum units have to satisfy \emph{smoothness}, the product units have to satisfy \emph{decomposability}, and the input units have to be \emph{tractable} \citep{choi2020probabilistic,vergari2021compositional}.

\paragraph{Permutation invariance.} Exchangeable data structures, including sets, graphs, partitions, and arrays \citep{orbanz2014bayesian}, have a factorial number of possible configurations (orderings). Permutation invariance says that no matter a selected configuration, the probability of an exchangeable data structure has to remain the same. To define the permutation invariance of a probability distribution, let $\mathbb{S}_n$ be a finite symmetric group of a set of $n$ elements. This is a set of all $n!$ permutations of $[n]\coloneqq(1,\ldots,n)$, where any of its members, $\pi\in\mathbb{S}_n$, permutes an $n$-dimensional vector, $X\coloneqq(X_1,\ldots,X_n)$, as follows: $\pi X=(X_{\pi(1)},\ldots,X_{\pi(n)})$. The probability distribution $p$ is permutation invariant iff $p(X)=p(\pi X)$ for all $\pi\in\mathbb{S}_n$. $X$ is permutation invariant if $p(X)$~is.

SPNs are only partially permutation invariant \citep{papez2024sum}, and so, for this paper, we consider SPNs as permutation-sensitive distributions.

\paragraph{Problem definition.} Let $G\coloneqq(X,A)$ be an $n$-node graph characterized by a feature matrix, $X\in\mathcal{X}^{n}$, and an adjacency tensor, $A\in\mathcal{A}^{n\times n}$. For example, to express $G$ with $q$ types of nodes and $r$ types of edges, we use one-hot encoding to define $\mathcal{X}\coloneqq(0,1)^q$ and $\mathcal{A}\coloneqq(0,1)^{r+1}$, where the extra category (+1) in $\mathcal{A}$ is to reflect the fact that there can be no connections between two nodes. Our objective is to learn a \emph{tractable} probability distribution over graphs, $p(G)$, given a collection of observed graphs, $(G_1,\ldots,G_N)$, where each $G_i$ has a different number of nodes and edges. We would like our GraphSPNs to inherit the ability to tractably answer the same inference quires as the conventional SPNs.

\section{Graph Sum-Product Networks}\label{sec:graphspn}
A graph sum-product network (GraphSPN) is a probability distribution over a graph, $p(G)$, where $G$ is random in its values and size. We conveniently use the fact that $G$ is represented by the feature matrix and the adjacency tensor, $G\coloneqq(X,A)$, and design the GraphSPN as a joint probability distribution of $X$ and~$A$,
\begin{equation}\label{eq:graphspn}
    p(G)
    \coloneqq
    g_n(p^{\text{spn}}_{m,n}(X,A))
    ,
\end{equation}
where $g_n$ is an operator ensuring the permutation invariance, and $p^{\text{spn}}_{m,n}$ is a permutation-sensitive SPN, as described in the next paragraph. The joint support of $p^{\text{spn}}_{m,n}$ leaves us with many choices to reorganize the elements of $X$ and $A$. We first concatenate $X$ and $A$ to form an $n$ by $n+1$ matrix and then vectorize this matrix in a row-wise manner (Figure~\ref{fig:padding}). In this way, there is always a node followed by its associated edges, $(X,A)\coloneqq(X_1,$ $A_1,$ $\ldots,$ $X_n,$ $A_n)$, where $X_i\in\mathcal{X}$ and $A_i\in\mathcal{A}^{n}$ are the $i$th slice of $X$ and $A$ along the first dimension, respectively.

\paragraph{Virtual node-padding.} SPNs assume fixed-size inputs (Section \ref{sec:background}). To deal with the varying size of $G$, we design the SPN in \eqref{eq:graphspn} as $p^{\text{spn}}_{m,n}(X,A)\coloneqq p^{\text{spn}}_m(\verb+padding+_{m,n}(X,A))$. It is a composition of a padding operator, $\verb+padding+_{m,n}$, and a permutation-sensitive SPN, $p^{\text{spn}}_m$, whose support (the root scope) has the size $m(m+1)$. $\verb+padding+_{m,n}$ extends $\mathcal{X}$ by an extra category to express that there is no node at a specific position in the graph. Consequently, each $n$-node instance of $G$ is filled in with virtual nodes (and associated virtual edges) to match the maximal number of nodes, $m$, which we expect in the training data (Figure~\ref{fig:padding}b). This technique---which we refer to as \emph{virtual node-padding}---is often used in deep generative models for graphs \citep{madhawa2019graphnvp}.


\paragraph{Exact permutation invariance.} The general way to design permutation invariant $p(G)$ is to make $g_n$ in \eqref{eq:graphspn} an average over all permutations $\pi\in\mathbb{S}_n$ of $(X,A)$,
\begin{equation}\label{eq:exact}
    p(G)
    \coloneqq
    \frac{1}{n!}\sum_{\pi\in\mathbb{S}_n}p^{\text{spn}}_{m,n}(\pi X,\pi A)
    ,
\end{equation}
where $\pi A\coloneqq(\pi A_{\pi_1},\ldots,$ $\pi A_{\pi_n})$.
This approach can be seen as artificially extending each instance of $(X,A)$ to $n!$ of its permutations. All the components $p^{\text{spn}}_{m,n}$ of the mixture \eqref{eq:exact} are an identical SPN, i.e., the components share the same parameterization.
This approach is computationally costly, as it requires $n!$ passes through $p^{\text{spn}}_{m,n}$, the scope of which has $m(m+1)$ entries. Note that \eqref{eq:exact} is known as the Janossy distribution \citep{daley2003introduction}.

\paragraph{Sorting.} Any graph $G$ has up to $n!$ equivalent permutations, $\pi G$, $\pi\in\mathbb{S}_n$. If we ensure that $p^{\text{spn}}_{m,n}$ always sees the same, user-defined, canonical representation of $G$, then the permutation invariance of \eqref{eq:graphspn} is guaranteed. One way to find such a representation is to \emph{sort} each instance of $G$ before entering $p^{\text{spn}}_{m,n}$. Then, composing a sorting function, \verb+sort+, with $p^{\text{spn}}_{m,n}$ avoids the summation over $n!$ terms in \eqref{eq:exact}, and thus circumvents the use of $g_n$, as follows:
\begin{equation}\label{eq:sorting}
    p(G)
    \coloneqq
    p^{\text{spn}}_{m,n}(\verb+sort+(X,A))
    ,
\end{equation}
where it holds that $\verb+sort+(\pi X,\pi A)\!=\!\verb+sort+(X,A)$ for all $\pi\in\mathbb{S}_n$. The computational complexity of \eqref{eq:sorting} depends on a specific sorting algorithm, e.g., $\mathcal{O}(n\log n)$. After the sorting, there is the need for only a single forward pass through $p^{\text{spn}}_{m,n}$, a significant reduction compared to \eqref{eq:exact}. However, certain orderings are more suitable than others \citep{montavon2012learning,niepert2016learning,defferrard2016convolutional}, and the right ordering often depends on domain knowledge.


\paragraph{Approximate permutation invariance.} The exact invariance \eqref{eq:exact} is computationally infeasible for all but small graphs. Indeed, even for $G$ with, say, 8 nodes, we have to perform 8! (40320) passes through $p^{\text{spn}}_{m,n}$, which is prohibitive when $p^{\text{spn}}_{m,n}$ is large. This permutation mechanism will slow down the inference as well. For example, marginalizing out certain node(s) has to be done 8! times. Therefore, we consider different ways to relax the exact permutation invariance, which have been used to design approximately permutation invariant neural networks, i.e., transformations $\mathbb{R}^u\rightarrow\mathbb{R}^v$ \citep{murphy2018janossy,murphy2019relational,wagstaff2022universal}. We cast them into their probabilistic interpretation, such that they are probability distributions $\mathbb{R}^u\rightarrow\mathbb{R}^+$.

\paragraph{$k$-ary sub-graphs.} The $k$-ary permutation invariance reduces the complexity of \eqref{eq:exact} by averaging over all $k$-node induced sub-graphs of $G$,
\begin{equation}\label{eq:kary}
    p(G)
    \colonapprox
    \frac{(n-k)!}{n!}\sum_{\mathbf{t}\in\mathbb{T}^k_n}p^{\text{spn}}_{k,n}(X_\mathbf{t}, A_{\mathbf{t}\mathbf{t}})
    .
\end{equation}
The $k$-node induced sub-graphs are represented by the feature sub-matrix, $X_\mathbf{t}\coloneqq(X_{t_1},\ldots,X_{t_k})\in\mathcal{X}^k$, and the adjacency sub-tensor, $A_{\mathbf{t}\mathbf{t}}\coloneqq(A_{t_1 t_1},\ldots,A_{t_k t_k})\in\mathcal{A}^{k\times k}$. They are selected by a set of $k$ indices, $\mathbf{t}\coloneqq(t_1,\ldots,t_k)\in\mathbb{T}^k_n$, where $\mathbb{T}^k_n$ is the set of all ways to choose $k$ elements out of $n$ unique elements (i.e., the set of all $k$-tuples of $[n]$). The main benefit of this method is that (i) the scope of $p^{\text{spn}}_{k,n}$ has a fixed size, $k(k+1)$, which also reduces the overall size of $p^{\text{spn}}_{k,n}$, and (ii) the number of passes through $p^{\text{spn}}_{k,n}$ is lower, $n!/(n-k)!<n!$. The expressivity of this approach grows with $k$. For $k=1$, we obtain a probabilistic version of Deep Sets \citep{zaheer2017deep}, whereas for $k=n$, we recover~\eqref{eq:exact}. For high $k$, we capture more dependencies among the nodes, whereas for low $k$, we sacrifice some of~them.

\paragraph{Random sampling.} The exact invariance \eqref{eq:exact} is a marginal distribution, $p(G)\coloneqq\sum_{\pi\in\mathbb{S}_n}p(\pi)p^{\text{spn}}_{m,n}(\pi X,\pi A)$, where the permutation $\pi$ is the latent variable, and $p(\pi)\coloneqq 1/n!$ is the uniform distribution on $\mathbb{S}_n$. This interpretation suggests to reduce the complexity of the full permutation invariance by approximating \eqref{eq:exact} with the random average,
\begin{equation}\label{eq:random}
    p(G)
    \colonapprox
    \frac{1}{N}\sum_{\pi\in\mathbb{S}^N_n}p^{\text{spn}}_{m,n}(\pi X,\pi A)
    .
\end{equation}
Here, $\mathbb{S}^N_n$ is a subset of $N<n!$ random permutations from the full set $\mathbb{S}_n$, which are sampled without repetition. The expressivity of \eqref{eq:random} depends on $N$. Even for $N=1$, we can still achieve sufficient approximate permutation invariance, assuming we draw a new sample at each training step and perform the training sufficiently long. For $N=n!$, we recover \eqref{eq:exact}. In the inference regime, it is important to use $N\gg1$. However, we can suffer from the risk of drawing less meaningful permutations for a given graph.

\section{Experiments}\label{sec:experiments}
We evaluate GraphSPNs in terms of their capacity to capture the underlying data distribution and their ability to generate realistic and novel graphs. We perform the experiments in the context of the computational design of molecular graphs. We provide the code at \url{https://github.com/mlnpapez/GraphSPN}.

\paragraph{Molecule generation.} Deep generative models for molecular graphs have recently gained significant attention. They hold great potential in important applications, such as discovering drugs and materials with desired chemical properties. Given a dataset of molecular graphs, the task is to learn a probability distribution of chemically valid molecules, $p(G)$. This is an intricate combinatorial problem, where not all combinations of atoms and bonds can be connected, but the connections must satisfy chemical valency constraints. We want $p(G)$ to generate molecules that were not seen during the training and satisfy the chemical valency~rules.

\paragraph{Dataset.} We test GraphSPNs on QM9 \citep{ramakrishnan2014quantum} dataset, a standard benchmark often used to assess deep generative models for molecular design. QM9 contains around 134k stable, small organic molecules with at most 9 atoms of 4 different types.

\begin{figure}
    \centering
    \vspace{-8pt}
    \includegraphics[width=0.9\linewidth]{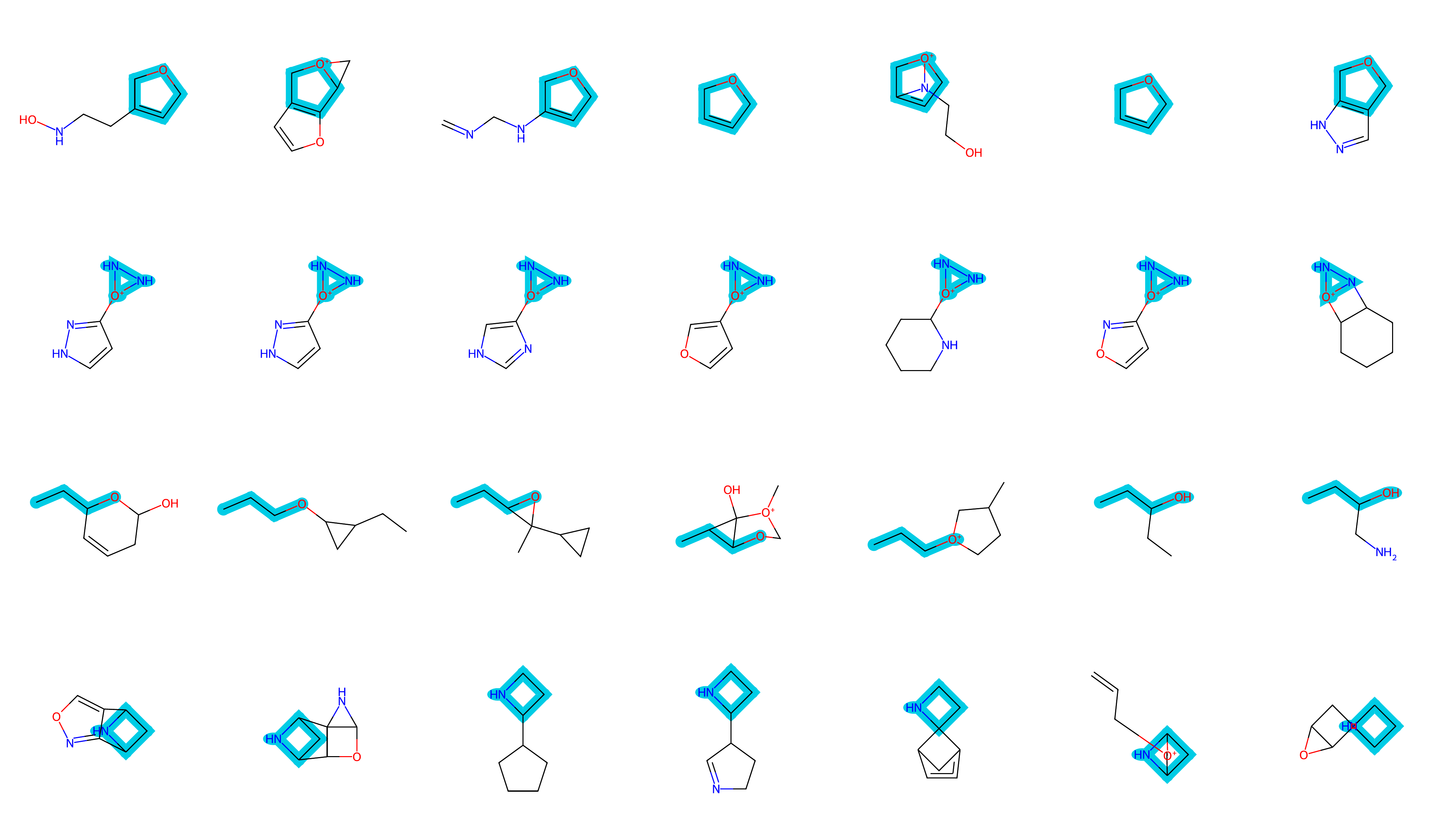}
    \vspace{-12pt}
    \caption{\emph{Conditional molecule generation on the QM9 dataset.} The blue area highlights the known part of the molecule. There is one such known part per row. Each column corresponds to a new molecule that is generated conditionally on the known part.}
    \label{fig:conditioning}
\end{figure}

\paragraph{Metrics.} We adopt the standard metrics for the molecule generation task. \emph{Validity} (V) and \emph{Validity w/o check} (V w/o check) are the percentages of chemically valid molecules (i.e., those not violating the chemical valency rules) in all generated molecules with and without any correction mechanisms, respectively. \emph{Uniqueness} (U) is the percentage of all generated molecules that are valid and unique (i.e., not a duplicate of some other generated molecule). \emph{Novelty} (N) is the percentage of valid molecules not in the training data.

\paragraph{GraphSPN variants.} We consider the following GraphSPN variants: \textsf{sort}, $k$\textsf{-ary}, and \textsf{rand} that implement \eqref{eq:sorting}, \eqref{eq:kary}, and \eqref{eq:random}, respectively, and the \textsf{none} variant with no invariance properties ($g_n$ in \eqref{eq:graphspn} is identity).

\begingroup
\setlength{\tabcolsep}{3pt}
\renewcommand{\arraystretch}{0.8}
\begin{table}[!t]
    \centering
    \small
    \begin{tabular}{lrrrr}
    \toprule
    Model & V & V w/o check & U & N \\
    \midrule
    GraphVAE & 55.70 & n/a & 76.00 & 61.60 \\
    GVAE & 60.20 & n/a & 9.30 & 80.90 \\
    CVAE & 10.20 & n/a & 67.50 & 90.00 \\
    RVAE & 96.60 & n/a & n/a & 95.50 \\
    GraphNVP & 83.10 & n/a & 99.20 & 58.20 \\
    GRF & 84.50 & n/a & 66.00 & 58.60 \\
    GraphAF & \color{red} \textbf{100.00} & 67.00 & 94.20 & 88.80 \\
    GraphDF & \color{red} \textbf{100.00} & 82.70 & 97.60 & 98.10 \\
    MoFlow & \color{red} \textbf{100.00} & \color{red} \textbf{89.00} & 98.50 & 96.40 \\
    ModFlow & n/a & 99.10 & 99.30 & \color{red} \textbf{100.00} \\
    \midrule
    GraphSPN: \textsf{none} & \color{red} \textbf{100.00} & 19.75 & \color{red} \textbf{100.00} & 99.49 \\
    GraphSPN: \textsf{rand} & \color{red} \textbf{100.00} & 15.38 & \color{red} \textbf{100.00} & 99.67 \\
    GraphSPN: \textsf{sort} & \color{red} \textbf{100.00} & 81.62 & 98.65 & 69.08 \\
    GraphSPN: $k$\textsf{-ary} & \color{red} \textbf{100.00} & 2.10 & \color{red} \textbf{100.00} & 98.30 \\
    \bottomrule
    \end{tabular}
    \vspace{-7pt}
    \caption{\emph{Unconditional molecule generation on the QM9 dataset.} The percentage of valid, valid w/o check, unique, and novel molecules for \emph{intractable}, permutation-sensitive models that all rely on the canonical ordering of atoms (above) and the \emph{tractable} GraphSPNs with various forms of permutation invariance (below).}
    \label{tab:metrics}
\end{table}
\endgroup

\paragraph{Results.} Table~\ref{tab:metrics} shows that all the GraphSPN variants achieve 100\% validity. This is because we use the posthoc validity correction mechanism from \citep{zang2020moflow} to satisfy the valency rules. However, the effectiveness of this correction depends on the validity w/o check. Naturally, if we have to correct more (possibly large) molecules, this quickly becomes computationally expensive. From this perspective, the \textsf{sort} variant delivers the best results, as its validity w/o check is highest. This result is because \textsf{sort} imposes the canonical ordering of atoms in each molecule, which is not satisfied by the other GraphSPN variants. $p^{\text{spn}}_{m,n}$ of the \textsf{none}, \textsf{rand}, and \textsf{sort} variants has the same architecture. Still, it is hard for the \textsf{none} and \textsf{rand} variants to capture the underlying data distribution. This result leads us to conclude that departing from the canonical ordering yields a more complicated distribution (with more modes), which is harder to capture sufficiently well, as evidenced by the poor ability to learn the essence of this problem, i.e., the valency rules. In the top of Table~\ref{tab:metrics}, we show baseline models (see Section~\ref{sec:details}). They are permutation-sensitive and also rely on the canonical ordering. We can see that the \textsf{sort} variant is competitive to---and sometimes even outperforms---these~baselines.

To illustrate that GraphSPNs provide tractable probabilistic inference, Figure~\ref{fig:conditioning} displays conditional molecule generation for the \textsf{sort} variant. Here, it can be seen that each molecule's newly generated part varies in size and composition.

\section{Conclusion}
The study of permutation invariance in the SPN community has received limited attention. This paper has the ambition to make a few steps towards improving upon this state. We have investigated different ways of ensuring the permutation invariance of SPNs and proposed GraphSPNs, a novel class of tractable deep generative models for exact and efficient probabilistic inference over graphs. We show that GraphSPNs are competitive with existing deep generative models for graphs. Among the proposed GraphSPN variants, the best performance was achieved with the one based on canonical ordering. This shows that the canonical ordering simplifies the target data distribution and makes it easier for an SPN---which is less densely connected than the models based on neural networks---to be trained successfully.

\begin{acknowledgements}
The authors acknowledge the support of the GA\v{C}R grant no.\ GA22-32620S and the OP VVV funded project CZ.02.1.01/0.0/0.0/16\_019/0000765 ``Research Center for Informatics''.
\end{acknowledgements}






\balance
\bibliography{tpm2024}

\newpage
\appendix
\onecolumn

\section{More notes on permutation invariance}

\begin{figure*}
    \centering
    \scalebox{0.67}{\input{principles.tikz}}
    \vspace{-20pt}
    \caption{\emph{An illustration of the key principles behind various GraphSPNs.} (a) The exact permutation invariance first computes $p^{\text{spn}}_{m,n}$ for all $n!$ permutations and then averages the results in \eqref{eq:exact}. (b) The $k$-ary permutation invariance approximates the exact invariance (a) by computing $p^{\text{spn}}_{k,n}$ (with the root scope of size $k(k+1)$) for all ways to choose $k$-node sub-graphs from the $n$-node original graph, $G$, without repetition and with the order, and then averaging the $M=n!/(n-k)!<n!$ results in \eqref{eq:kary}. (c) The random sampling approximates the exact invariance (a) by computing the average only for $N<n!$ permutations of $G$ in \eqref{eq:random}. (d) The sorting approach is also exact, but it simplifies the target data distribution by first imposing the same canonical ordering of $G$ and then computing $p^{\text{spn}}_{m,n}$, as displayed in \eqref{eq:sorting}.}
    \label{fig:principles}
\end{figure*}
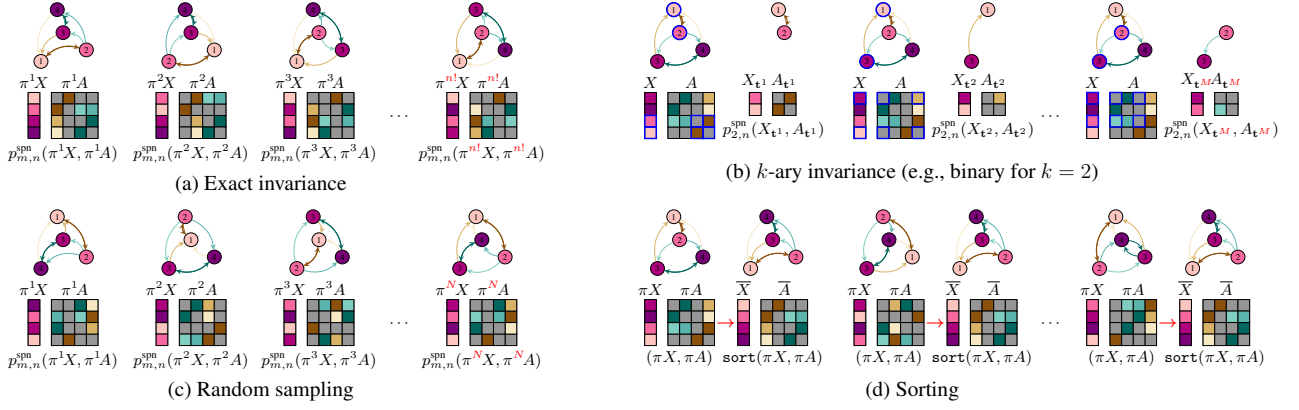

Figure~\ref{fig:principles} illustrates the fundamental principles underlying the GraphSPN variants discussed in Section~\ref{sec:graphspn}.

\section{Tractability}



\paragraph{Tractability.} An SPN is tractable if it answers probabilistic queries \emph{exactly} and \emph{efficiently}. Here, exact means that the answers do not involve any approximation or heuristic, and efficient means that the answers can be obtained (computed) in polynomial time \citep{choi2020probabilistic}. We are interested in probabilistic queries that can collectively be expressed in terms of the following expectation:
\begin{equation}\label{eq:expectation}
    \nu(f)
    \coloneqq
    \int f(G)p(G)dG
    ,
\end{equation}
where $f(G)$ is a function that allows us to formulate a desired query over (a part of) $G$. We defer concrete examples of $f(G)$ to Section~\ref{sec:queries}. The expectation \eqref{eq:expectation} admits a closed-form solution only if both $f$ and $p$ satisfy certain structural constraints.

\begin{assumption}{(Constraints on $p$).}\label{ass:p-constraints}
    We consider the following constraints. (i) \textbf{Smoothness}: each $u\in\mathsf{L}_\mathsf{S}$ satisfies $\forall a,b\in\mathbf{in}(u):\psi_a=\psi_b$, where $\mathbf{in}(u)$ is the set of inputs of a $u$-unit. (ii) \textbf{Decomposability}: each $u\in\mathsf{L}_\mathsf{P}$ satisfies $\forall a,b\in\mathbf{in}(u):\psi_a\cap\psi_b=\varnothing$. (iii) \textbf{Tractable input layers}: $\int f_u(\psi_u)_up(\psi_u)d\psi_u$ is tractable for each $u\in\mathsf{L}_\mathsf{I}$.
\end{assumption}

Assumption \ref{ass:p-constraints}(iii) requires that each $p_i(\psi)$ is from a tractable family of probability distributions \citep{barndorff1978information} and $f_i(\psi)$ is such that the integral admits an algebraically closed-form solution.

\begin{assumption}{(Constraints on $f$).}\label{ass:f-constraints}
   The function $f(G)\coloneqq f(X_1,A_1\ldots,X_n,A_n)$ is \emph{omni-compatible} \citep{vergari2021compositional} with respect to $p^{\text{spn}}_{m,n}$.
\end{assumption}

\begin{proposition}{(Tractability of GraphSPNs.)}\label{prop:tractability}
Let $p(G)$ be a GraphSPN \eqref{eq:graphspn} satisfying Assumptions \ref{ass:p-constraints}, and let $f(G)$ be a function satisfying Assumption \ref{ass:f-constraints}. Then, the integral \eqref{eq:expectation} is tractable.
\end{proposition}

Proposition \ref{prop:tractability} covers all the GraphSPN variants \eqref{eq:exact}, \eqref{eq:sorting}, and \eqref{eq:random}, except the $k$-ary sub-graphs \eqref{eq:kary}. Indeed, the average in \eqref{eq:exact} and \eqref{eq:random} can be seen as a sum unit with fixed uniform weights, which satisfies the smoothness assumption. The sorting operation in \eqref{eq:sorting} does not affect the tractability.
However, regarding the $k$-ary GraphSPNs, the average in \eqref{eq:kary} does not satisfy the smoothness assumption (i.e., it is not a smooth sum unit), as each of its children has a different scope. To make the $k$-ary variant tractable, we must perform the smoothing \citep{shih2019smoothing,choi2020probabilistic}.

\subsection{Tractable inference queries over graphs}\label{sec:queries}

\paragraph{Querying $r$-node sub-graphs.}

An omni-compatible function $f$ (Assumption \ref{ass:f-constraints}) that targets an $r$-node sub-graph of $G$ can be expressed as follows:
\begin{equation}\label{eq:f}
    f(X,A)\coloneqq
    \prod_{i\in\mathbf{a}}h(X_i)
        \left(
        \prod_{j\in\mathbf{a}}h(A_{ij})
        \prod_{k\notin\mathbf{a}}h(A_{ik})
        \right)
    \prod_{t\notin\mathbf{a}}h(X_t)
        \left(
        \prod_{u\notin\mathbf{a}}h(A_{tu})
        \prod_{v\in\mathbf{a}}h(A_{tv})
        \right),   
\end{equation}
where $\mathbf{a}\coloneqq\{a_1,\ldots,a_r\} \subseteq \{1, \hdots, n\}$ specifies indices of $r$ nodes presented in a queried sub-graph.

\paragraph{Marginalizing selected nodes and (or) edges.} The marginal query is realized by an appropriate choice of $h$ for $X_i$ and $A_{ij}$. The choice $h(X_i)\coloneqq\mathds{1}_\mathcal{B}(X_i)$, where $\mathds{1}_\mathcal{B}$ is the indicator function and $\mathcal{B}\coloneqq\mathcal{X}$, corresponds to marginalizing node feature of $i$-th node in the graph $G$. The evidence query is obtained for $\mathcal{B}\coloneqq X_i$. Similarly, choosing $h(A_{ij})\coloneqq\mathds{1}_\mathcal{B}(A_{ij})$ and $\mathcal{B}\coloneqq\mathcal{A}$ corresponds to marginalizing edge between nodes $i$ and $j$ in the graph $G$. Again, the evidence query is given by setting $\mathcal{B}\coloneqq A_{ij}$. We demonstrate an example of the marginal query for $\mathbf{a}\coloneqq\{2\}$ in Figure~\ref{fig:principles}.

\begin{figure*}
    \centering
    \scalebox{0.67}{\input{inference.tikz}}
    \vspace{-10pt}
    \caption{\emph{An example of a tractable inference query over a graph.} (a) An instantiation of the omni-compatible function \eqref{eq:f} over a graph $G$ for $\mathbf{a}\coloneqq\{2\}$. The blue color highlights the targeted node and its associated edges. (b) A visual and matrix representation of $G$, where the targeted node and its associated edges, which correspond to (a), are highlighted in blue. (c) After targeting the node and its connected edges, we can perform, e.g., the marginal query and obtain a marginal graph $\overline{G}$.}
    \label{fig:inference}
\end{figure*}

%









\section{Experimental Details}\label{sec:details}

\paragraph{Preprocessing.} We use the RDKit library \citep{landrum2006rdkit} to first kekulize the molecules and then remove the hydrogen atoms. The final molecules contain only the single, double, and triple bonds. Since we aim to test the permutation invariance of various GraphSPNs, we randomly permute the atoms in each molecule during the preprocessing.

\paragraph{Baselines.} We compare GraphSPNs with various molecular deep generative models: grammar variational autoencoder (GVAE) \citep{kusner2017grammar}, character VAE (CVAE) \citep{gomez2018automatic}, regularizing VAE (RVAE) \citep{ma2018constrained}, junction tree VAE (JT-VAE) \citep{jin2018junction}, graph VAE (GraphVAE) \citep{simonovsky2018graphvae}, graph real-valued non-volume preserving (GraphNVP) flow \citep{madhawa2019graphnvp}, graph autoregressive flow (GraphAF) \citep{shi2020graphaf}, graph discrete flow (GraphDF) \citep{luo2021graphdf}, molecular flow (MoFlow) \citep{zang2020moflow}, modular flow (ModFlow) \citep{verma2022modular}, and graph residual flow (GRF) \citep{honda2019graph}.

\paragraph{Architecture.} To implement the permutation-sensitive part of GraphSPNs, we adopt the EinSum networks \citep{peharz2020einsum}. The expressive power of this monolithic and tensorized variant of SPNs is driven by four hyper-parameters: the number of layers $n_l\in\{1, 2, 3\}$, the number of sum units $n_S\in\{10, 40, 80\}$, the number of input units $n_I\in\{10, 40\}$, and the number of repetitions $n_R\in\{10, 40, 80\}$. The input units are categorical distributions. Molecules are undirected acyclic graphs. To satisfy this constraint, we use only the lower triangular part of $A$ in the sampling procedure.

For the \textsf{rand} variant of GraphSPNs \eqref{eq:random}, we use $\mathbb{S}_m^{20}$, i.e., 20 permutations. For the $k$\textsf{-ary} variant \eqref{eq:kary}, we use $k=2$ to obtain the average with only 72 elements. To impose the canonical ordering with the \textsf{sort} variant \eqref{eq:sorting}, we use the RDKit library.

\paragraph{Learning.} The results for the baseline methods in the top part of Table \ref{tab:metrics} are obtained from their respective papers. For all the GraphSPN variants in the bottom part of Table \ref{tab:metrics}, we minimize negative log-likelihood for $40$ epochs using the ADAM optimizer \citep{kingma2014adam} with $256$
samples in the minibatch, step-size $\alpha=0.05$ and decay rates $\beta_1=0.9$ and $\beta_2=0.82$. All experiments are repeated $5$ times with different initialization of the model's parameters. We sample 4000 molecules to compute the metrics introduced in Section \ref{sec:experiments}.

\section{Additional results}\label{sec:additional_results}
\begin{figure}
    \centering
    \includegraphics[width=0.9\linewidth]{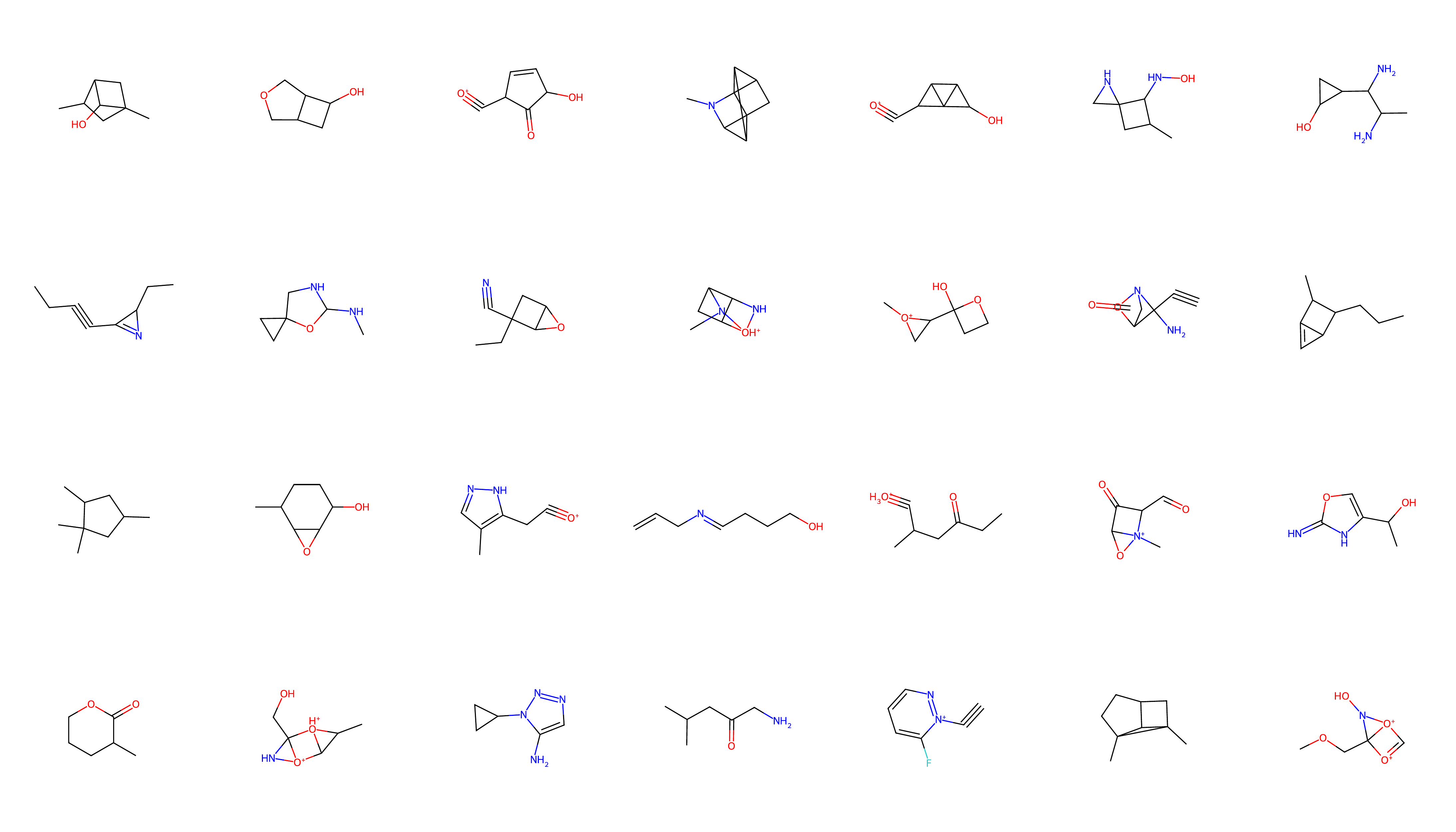}
    \caption{\emph{Unconditional molecule generation on the QM9 dataset.}}
    \label{fig:generation}
\end{figure}

Figure~\ref{fig:generation} shows unconditional samples of molecules from the \textsf{sort} variant \eqref{eq:sorting} that was trained on the QM9 dataset. The resulting molecules resemble those from the training data, yet they are all newly discovered molecules.

\end{document}

%% file: padding.tikz
\newcommand\grid{6.5pt}

\begin{tikzpicture}[font=\fontsize{5}{3}\selectfont,x=\grid,y=\grid]

\definecolor{c0}{RGB}{255,255,255}
\definecolor{c1}{RGB}{27,158,119}
\definecolor{c2}{RGB}{117,112,179}
\definecolor{c3}{RGB}{217,95,2}
\definecolor{c4}{RGB}{231,41,138}
\definecolor{c5}{RGB}{150,150,150}

\coordinate (a)  at (-21,0);
\coordinate (a1) at ($(a) + (-5.0,+0.0)$);
\coordinate (a2) at ($(a) + (-2.0,+0.0)$);
\coordinate (b)  at (+13.3,5);
\coordinate (b1) at ($(b) + (-24.0,+2.0)$);
\coordinate (b2) at ($(b) + (-24.0,+0.0)$);
\coordinate (c)  at (+12.3,-1);
\coordinate (c1) at ($(c) + (-24.0,+2.0)$);
\coordinate (c2) at ($(c) + (-24.0,+0.0)$);
\coordinate (c3) at ($(c) + (-24.0,-2.0)$);

\node at ($(a) + (0.0,-1.5)$) {\tiny (a) Graph representation};

\draw[step=\grid, black] ($(a1) + (0,0)$) grid ($(a1) + (1,8)$) node[label={[xshift=-0.5*\grid, yshift=-0.8*\grid]$X$}] {};

\draw[draw=black,fill=c0] ($(a1) + (0,1)$) rectangle ($(a1) + (1,3)$) node[xshift=-0.5*\grid, yshift=-0.6*\grid] {$\vdots$};
\draw[draw=black,fill=c1] ($(a1) + (0,4)$) rectangle ($(a1) + (1,5)$) node[xshift=-0.5*\grid, yshift=-0.6*\grid] {};
\draw[draw=black,fill=c1] ($(a1) + (0,5)$) rectangle ($(a1) + (1,7)$) node[xshift=-0.5*\grid, yshift=-0.6*\grid] {$\vdots$};
\draw[draw=black,fill=c1] ($(a1) + (0,7)$) rectangle ($(a1) + (1,8)$) node[xshift=-0.5*\grid, yshift=-0.6*\grid] {};

\draw ($(a1) + (-0.9*\grid,7.5)$) node[label={[xshift=0.5*\grid, yshift=-1.35*\grid]$1$}] {};
\draw ($(a1) + (-0.9*\grid,4.5)$) node[label={[xshift=0.4*\grid, yshift=-1.35*\grid]$n$}] {};
\draw ($(a1) + (-0.9*\grid,3.5)$) node[label={[xshift=0.0*\grid, yshift=-1.35*\grid]$n\texttt{+}1$}] {};
\draw ($(a1) + (-0.9*\grid,0.5)$) node[label={[xshift=0.3*\grid, yshift=-1.35*\grid]$m$}] {};

\draw[step=\grid, black] ($(a2) + (0,0)$) grid ($(a2) + (8,8)$) node[label={[xshift=-4.0*\grid, yshift=-0.8*\grid]$A$}] {};

\draw[draw=black,fill=c0] ($(a2) + (0,1)$) rectangle ($(a2) + (1,3)$) node[xshift=-0.5*\grid, yshift=-0.6*\grid] {$\vdots$};
\draw[draw=black,fill=c0] ($(a2) + (3,1)$) rectangle ($(a2) + (4,3)$) node[xshift=-0.5*\grid, yshift=-0.6*\grid] {$\vdots$};
\draw[draw=black,fill=c0] ($(a2) + (4,1)$) rectangle ($(a2) + (5,3)$) node[xshift=-0.5*\grid, yshift=-0.6*\grid] {$\vdots$};
\draw[draw=black,fill=c0] ($(a2) + (7,1)$) rectangle ($(a2) + (8,3)$) node[xshift=-0.5*\grid, yshift=-0.6*\grid] {$\vdots$};
\draw[draw=black,fill=c0] ($(a2) + (4,5)$) rectangle ($(a2) + (5,7)$) node[xshift=-0.5*\grid, yshift=-0.6*\grid] {$\vdots$};
\draw[draw=black,fill=c0] ($(a2) + (7,5)$) rectangle ($(a2) + (8,7)$) node[xshift=-0.5*\grid, yshift=-0.6*\grid] {$\vdots$};
\draw[draw=black,fill=c0] ($(a2) + (1,0)$) rectangle ($(a2) + (3,1)$) node[xshift=-1.0*\grid, yshift=-0.5*\grid] {$\ldots$};
\draw[draw=black,fill=c0] ($(a2) + (1,3)$) rectangle ($(a2) + (3,4)$) node[xshift=-1.0*\grid, yshift=-0.5*\grid] {$\ldots$};
\draw[draw=black,fill=c0] ($(a2) + (5,0)$) rectangle ($(a2) + (7,1)$) node[xshift=-1.0*\grid, yshift=-0.5*\grid] {$\ldots$};
\draw[draw=black,fill=c0] ($(a2) + (5,3)$) rectangle ($(a2) + (7,4)$) node[xshift=-1.0*\grid, yshift=-0.5*\grid] {$\ldots$};
\draw[draw=black,fill=c0] ($(a2) + (5,3)$) rectangle ($(a2) + (7,4)$) node[xshift=-1.0*\grid, yshift=-0.5*\grid] {$\ldots$};
\draw[draw=black,fill=c0] ($(a2) + (5,7)$) rectangle ($(a2) + (7,8)$) node[xshift=-1.0*\grid, yshift=-0.5*\grid] {$\ldots$};
\draw[draw=black,fill=c0] ($(a2) + (5,4)$) rectangle ($(a2) + (7,5)$) node[xshift=-1.0*\grid, yshift=-0.5*\grid] {$\ldots$};
\draw[draw=black,fill=c0] ($(a2) + (1,1)$) rectangle ($(a2) + (3,3)$) node[xshift=-1.0*\grid, yshift=-0.6*\grid] {$\ddots$};
\draw[draw=black,fill=c0] ($(a2) + (5,1)$) rectangle ($(a2) + (7,3)$) node[xshift=-1.0*\grid, yshift=-0.6*\grid] {$\ddots$};
\draw[draw=black,fill=c0] ($(a2) + (5,5)$) rectangle ($(a2) + (7,7)$) node[xshift=-1.0*\grid, yshift=-0.6*\grid] {$\ddots$};

\draw[draw=black,fill=c2] ($(a2) + (0,4)$) rectangle ($(a2) + (1,5)$) node[xshift=-0.5*\grid, yshift=-0.6*\grid] {};
\draw[draw=black,fill=c2] ($(a2) + (0,7)$) rectangle ($(a2) + (1,8)$) node[xshift=-0.5*\grid, yshift=-0.6*\grid] {};
\draw[draw=black,fill=c2] ($(a2) + (3,4)$) rectangle ($(a2) + (4,5)$) node[xshift=-0.5*\grid, yshift=-0.6*\grid] {};
\draw[draw=black,fill=c2] ($(a2) + (3,7)$) rectangle ($(a2) + (4,8)$) node[xshift=-0.5*\grid, yshift=-0.6*\grid] {};
\draw[draw=black,fill=c2] ($(a2) + (0,5)$) rectangle ($(a2) + (1,7)$) node[xshift=-0.5*\grid, yshift=-0.6*\grid] {$\vdots$};
\draw[draw=black,fill=c2] ($(a2) + (3,5)$) rectangle ($(a2) + (4,7)$) node[xshift=-0.5*\grid, yshift=-0.6*\grid] {$\vdots$};
\draw[draw=black,fill=c2] ($(a2) + (1,7)$) rectangle ($(a2) + (3,8)$) node[xshift=-1.0*\grid, yshift=-0.5*\grid] {$\ldots$};
\draw[draw=black,fill=c2] ($(a2) + (1,4)$) rectangle ($(a2) + (3,5)$) node[xshift=-1.0*\grid, yshift=-0.5*\grid] {$\ldots$};
\draw[draw=black,fill=c2] ($(a2) + (1,5)$) rectangle ($(a2) + (3,7)$) node[xshift=-1.0*\grid, yshift=-0.6*\grid] {$\ddots$};

\draw ($(a2) + (-0.9*\grid,7.5)$) node[label={[xshift=0.5*\grid, yshift=-1.35*\grid]$1$}] {};
\draw ($(a2) + (-0.9*\grid,4.5)$) node[label={[xshift=0.4*\grid, yshift=-1.35*\grid]$n$}] {};
\draw ($(a2) + (-0.9*\grid,3.5)$) node[label={[xshift=0.0*\grid, yshift=-1.35*\grid]$n\texttt{+}1$}] {};
\draw ($(a2) + (-0.9*\grid,0.5)$) node[label={[xshift=0.3*\grid, yshift=-1.35*\grid]$m$}] {};

\draw ($(a2) + (0.5,-0.4*\grid)$) node[label={[xshift=0.0*\grid, yshift=-1.4*\grid]$1$}] {};
\draw ($(a2) + (3.5,-0.4*\grid)$) node[label={[xshift=0.0*\grid, yshift=-1.4*\grid]$n$}] {};
\draw ($(a2) + (4.5,-0.4*\grid)$) node[label={[xshift=0.3*\grid, yshift=-1.4*\grid]$n\texttt{+}1$}] {};
\draw ($(a2) + (7.5,-0.4*\grid)$) node[label={[xshift=0.0*\grid, yshift=-1.4*\grid]$m$}] {};

\node at ($(b) + (0.0,-1.0)$) {\tiny (b) Virtual node-padding};

\draw[draw=black,fill=c1] ($(b1) + ( 0,0)$) rectangle ($(b1) + ( 1,1)$) node[label={[xshift=-0.5*\grid, yshift=-0.8*\grid]$X_1$}] {};
\draw[draw=black,fill=c2] ($(b1) + ( 1,0)$) rectangle ($(b1) + ( 2,1)$) node[xshift=-0.5*\grid, yshift=-0.6*\grid] {};
\draw[draw=black,fill=c2] ($(b1) + ( 2,0)$) rectangle ($(b1) + ( 4,1)$) node[xshift=-1.0*\grid, yshift=-0.5*\grid] {$\ldots$};
\draw[draw=black,fill=c2] ($(b1) + ( 4,0)$) rectangle ($(b1) + ( 5,1)$) node[label={[xshift=-0.0*\grid, yshift=-0.8*\grid]$A_1$}] {};
\draw[draw=black,fill=c0] ($(b1) + ( 5,0)$) rectangle ($(b1) + ( 6,1)$) node[xshift=-0.5*\grid, yshift=-0.6*\grid] {};
\draw[draw=black,fill=c0] ($(b1) + ( 6,0)$) rectangle ($(b1) + ( 8,1)$) node[xshift=-1.0*\grid, yshift=-0.5*\grid] {$\ldots$};
\draw[draw=black,fill=c0] ($(b1) + ( 8,0)$) rectangle ($(b1) + ( 9,1)$) node[xshift=-0.5*\grid, yshift=-0.6*\grid] {};

\draw[draw=none,fill=c0] ($(b1) + ( 9,0)$) rectangle ($(b1) + (12,1)$) node[xshift=-1.5*\grid, yshift=-0.5*\grid] {$\ldots$};

\draw[draw=black,fill=c1] ($(b1) + (12,0)$) rectangle ($(b1) + (13,1)$) node[label={[xshift=-0.5*\grid, yshift=-0.8*\grid]$X_n$}] {};
\draw[draw=black,fill=c2] ($(b1) + (13,0)$) rectangle ($(b1) + (14,1)$) node[xshift=-0.5*\grid, yshift=-0.6*\grid] {};
\draw[draw=black,fill=c2] ($(b1) + (14,0)$) rectangle ($(b1) + (16,1)$) node[xshift=-1.0*\grid, yshift=-0.5*\grid] {$\ldots$};
\draw[draw=black,fill=c2] ($(b1) + (16,0)$) rectangle ($(b1) + (17,1)$) node[label={[xshift=-0.0*\grid, yshift=-0.8*\grid]$A_n$}] {};
\draw[draw=black,fill=c0] ($(b1) + (17,0)$) rectangle ($(b1) + (18,1)$) node[xshift=-0.5*\grid, yshift=-0.6*\grid] {};
\draw[draw=black,fill=c0] ($(b1) + (18,0)$) rectangle ($(b1) + (20,1)$) node[xshift=-1.0*\grid, yshift=-0.5*\grid] {$\ldots$};
\draw[draw=black,fill=c0] ($(b1) + (20,0)$) rectangle ($(b1) + (21,1)$) node[xshift=-0.5*\grid, yshift=-0.6*\grid] {};

\draw[draw=none,fill=c0] ($(b1) + (21,0)$) rectangle ($(b1) + (24,1)$) node[xshift=-1.5*\grid, yshift=-0.5*\grid] {$\ldots$};

\draw[draw=black,fill=c0] ($(b1) + (24,0)$) rectangle ($(b1) + (25,1)$) node[label={[xshift=-0.5*\grid, yshift=-0.8*\grid]$X_{n\texttt{+}1}$}] {};
\draw[draw=black,fill=c0] ($(b1) + (25,0)$) rectangle ($(b1) + (26,1)$) node[xshift=-0.5*\grid, yshift=-0.6*\grid] {};
\draw[draw=black,fill=c0] ($(b1) + (26,0)$) rectangle ($(b1) + (28,1)$) node[xshift=-1.0*\grid, yshift=-0.5*\grid] {$\ldots$};
\draw[draw=black,fill=c0] ($(b1) + (28,0)$) rectangle ($(b1) + (29,1)$) node[label={[xshift=-0.1*\grid, yshift=-0.8*\grid]$A_{n\texttt{+}1}$}] {};
\draw[draw=black,fill=c0] ($(b1) + (29,0)$) rectangle ($(b1) + (30,1)$) node[xshift=-0.5*\grid, yshift=-0.6*\grid] {};
\draw[draw=black,fill=c0] ($(b1) + (30,0)$) rectangle ($(b1) + (32,1)$) node[xshift=-1.0*\grid, yshift=-0.5*\grid] {$\ldots$};
\draw[draw=black,fill=c0] ($(b1) + (32,0)$) rectangle ($(b1) + (33,1)$) node[xshift=-0.5*\grid, yshift=-0.6*\grid] {};

\draw[draw=none,fill=c0] ($(b1) + (33,0)$) rectangle ($(b1) + (36,1)$) node[xshift=-1.5*\grid, yshift=-0.5*\grid] {$\ldots$};

\draw[draw=black,fill=c0] ($(b1) + (36,0)$) rectangle ($(b1) + (37,1)$) node[label={[xshift=-0.5*\grid, yshift=-0.8*\grid]$X_m$}] {};
\draw[draw=black,fill=c0] ($(b1) + (37,0)$) rectangle ($(b1) + (38,1)$) node[xshift=-0.5*\grid, yshift=-0.6*\grid] {};
\draw[draw=black,fill=c0] ($(b1) + (38,0)$) rectangle ($(b1) + (40,1)$) node[xshift=-1.0*\grid, yshift=-0.5*\grid] {$\ldots$};
\draw[draw=black,fill=c0] ($(b1) + (40,0)$) rectangle ($(b1) + (41,1)$) node[label={[xshift=-0.0*\grid, yshift=-0.8*\grid]$A_m$}] {};
\draw[draw=black,fill=c0] ($(b1) + (41,0)$) rectangle ($(b1) + (42,1)$) node[xshift=-0.5*\grid, yshift=-0.6*\grid] {};
\draw[draw=black,fill=c0] ($(b1) + (42,0)$) rectangle ($(b1) + (44,1)$) node[xshift=-1.0*\grid, yshift=-0.5*\grid] {$\ldots$};
\draw[draw=black,fill=c0] ($(b1) + (44,0)$) rectangle ($(b1) + (45,1)$) node[xshift=-0.5*\grid, yshift=-0.6*\grid] {};

\node[draw=none] at ($(b2) + (-1.5,0.5)$) {$p^{\text{spn}}_{\textcolor{red}{m},n}$};
\draw[draw=black,fill=c1] ($(b2) + ( 0,0)$) rectangle ($(b2) + ( 1,1)$) node[label={[xshift=-0.5*\grid, yshift=-2.9*\grid]$1$}] {};
\draw[draw=black,fill=c2] ($(b2) + ( 1,0)$) rectangle ($(b2) + ( 2,1)$) node[label={[xshift=-0.5*\grid, yshift=-2.9*\grid]$2$}] {};
\draw[draw=black,fill=c2] ($(b2) + ( 2,0)$) rectangle ($(b2) + ( 4,1)$) node[xshift=-1.0*\grid, yshift=-0.5*\grid] {$\ldots$};
\draw[draw=black,fill=c2] ($(b2) + ( 4,0)$) rectangle ($(b2) + ( 5,1)$) node[xshift=-0.5*\grid, yshift=-0.6*\grid] {};
\draw[draw=black,fill=c5] ($(b2) + ( 5,0)$) rectangle ($(b2) + ( 6,1)$) node[xshift=-0.5*\grid, yshift=-0.6*\grid] {};
\draw[draw=black,fill=c5] ($(b2) + ( 6,0)$) rectangle ($(b2) + ( 8,1)$) node[xshift=-1.0*\grid, yshift=-0.5*\grid] {$\ldots$};
\draw[draw=black,fill=c5] ($(b2) + ( 8,0)$) rectangle ($(b2) + ( 9,1)$) node[xshift=-0.5*\grid, yshift=-0.6*\grid] {};

\draw[draw=black,fill=c0] ($(b2) + ( 9,0)$) rectangle ($(b2) + (12,1)$) node[xshift=-1.5*\grid, yshift=-0.5*\grid] {$\ldots$};

\draw[draw=black,fill=c1] ($(b2) + (12,0)$) rectangle ($(b2) + (13,1)$) node[xshift=-0.5*\grid, yshift=-0.6*\grid] {};
\draw[draw=black,fill=c2] ($(b2) + (13,0)$) rectangle ($(b2) + (14,1)$) node[xshift=-0.5*\grid, yshift=-0.6*\grid] {};
\draw[draw=black,fill=c2] ($(b2) + (14,0)$) rectangle ($(b2) + (16,1)$) node[xshift=-1.0*\grid, yshift=-0.5*\grid] {$\ldots$};
\draw[draw=black,fill=c2] ($(b2) + (16,0)$) rectangle ($(b2) + (17,1)$) node[xshift=-0.5*\grid, yshift=-0.6*\grid] {};
\draw[draw=black,fill=c5] ($(b2) + (17,0)$) rectangle ($(b2) + (18,1)$) node[xshift=-0.5*\grid, yshift=-0.6*\grid] {};
\draw[draw=black,fill=c5] ($(b2) + (18,0)$) rectangle ($(b2) + (20,1)$) node[xshift=-1.0*\grid, yshift=-0.5*\grid] {$\ldots$};
\draw[draw=black,fill=c5] ($(b2) + (20,0)$) rectangle ($(b2) + (21,1)$) node[xshift=-0.5*\grid, yshift=-0.6*\grid] {};

\draw[draw=black,fill=c0] ($(b2) + (21,0)$) rectangle ($(b2) + (24,1)$) node[xshift=-1.5*\grid, yshift=-0.5*\grid] {$\ldots$};

\draw[draw=black,fill=c5] ($(b2) + (24,0)$) rectangle ($(b2) + (25,1)$) node[xshift=-0.5*\grid, yshift=-0.6*\grid] {};
\draw[draw=black,fill=c5] ($(b2) + (25,0)$) rectangle ($(b2) + (26,1)$) node[xshift=-0.5*\grid, yshift=-0.6*\grid] {};
\draw[draw=black,fill=c5] ($(b2) + (26,0)$) rectangle ($(b2) + (28,1)$) node[xshift=-1.0*\grid, yshift=-0.5*\grid] {$\ldots$};
\draw[draw=black,fill=c5] ($(b2) + (28,0)$) rectangle ($(b2) + (29,1)$) node[xshift=-0.5*\grid, yshift=-0.6*\grid] {};
\draw[draw=black,fill=c5] ($(b2) + (29,0)$) rectangle ($(b2) + (30,1)$) node[xshift=-0.5*\grid, yshift=-0.6*\grid] {};
\draw[draw=black,fill=c5] ($(b2) + (30,0)$) rectangle ($(b2) + (32,1)$) node[xshift=-1.0*\grid, yshift=-0.5*\grid] {$\ldots$};
\draw[draw=black,fill=c5] ($(b2) + (32,0)$) rectangle ($(b2) + (33,1)$) node[xshift=-0.5*\grid, yshift=-0.6*\grid] {};

\draw[draw=black,fill=c0] ($(b2) + (33,0)$) rectangle ($(b2) + (36,1)$) node[xshift=-1.5*\grid, yshift=-0.5*\grid] {$\ldots$};

\draw[draw=black,fill=c5] ($(b2) + (36,0)$) rectangle ($(b2) + (37,1)$) node[xshift=-0.5*\grid, yshift=-0.6*\grid] {};
\draw[draw=black,fill=c5] ($(b2) + (37,0)$) rectangle ($(b2) + (38,1)$) node[xshift=-0.5*\grid, yshift=-0.6*\grid] {};
\draw[draw=black,fill=c5] ($(b2) + (38,0)$) rectangle ($(b2) + (40,1)$) node[xshift=-1.0*\grid, yshift=-0.5*\grid] {$\ldots$};
\draw[draw=black,fill=c5] ($(b2) + (40,0)$) rectangle ($(b2) + (41,1)$) node[xshift=-0.5*\grid, yshift=-0.6*\grid] {};
\draw[draw=black,fill=c5] ($(b2) + (41,0)$) rectangle ($(b2) + (42,1)$) node[xshift=-0.5*\grid, yshift=-0.6*\grid] {};
\draw[draw=black,fill=c5] ($(b2) + (42,0)$) rectangle ($(b2) + (44,1)$) node[xshift=-1.0*\grid, yshift=-0.5*\grid] {$\ldots$};
\draw[draw=black,fill=c5] ($(b2) + (44,0)$) rectangle ($(b2) + (45,1)$) node[label={[xshift=-1.0*\grid, yshift=-3.0*\grid]$\textcolor{red}{m(m\texttt{+}1)}$}] {};

\end{tikzpicture}

%% file: principles.tikz
\newcommand\grid{6.5pt}

\definecolor{c0}{RGB}{243,246,249}
\definecolor{c1}{RGB}{27,158,119}
\definecolor{c2}{RGB}{117,112,179}
\definecolor{c3}{RGB}{217,95,2}
\definecolor{c4}{RGB}{231,41,138}
\definecolor{c5}{RGB}{150,150,150}

\definecolor{n1}{RGB}{252,197,192}
\definecolor{n2}{RGB}{247,104,161}
\definecolor{n3}{RGB}{174,1,126}
\definecolor{n4}{RGB}{122,1,119}

\definecolor{e12}{RGB}{140,81,10}
\definecolor{e13}{RGB}{150,150,150}
\definecolor{e14}{RGB}{150,150,150}
\definecolor{e23}{RGB}{128,205,193}
\definecolor{e24}{RGB}{90,180,172}
\definecolor{e34}{RGB}{1,102,94}

\definecolor{e21}{RGB}{140,81,10}
\definecolor{e31}{RGB}{216,179,101}
\definecolor{e41}{RGB}{246,232,195}
\definecolor{e32}{RGB}{150,150,150}
\definecolor{e42}{RGB}{150,150,150}
\definecolor{e43}{RGB}{1,102,94}

\definecolor{c00}{RGB}{150,150,150}

\newcommand\va{0}
\newcommand\vb{1}
\newcommand\vc{2}
\newcommand\vd{3}

\newcommand{\nodes}[5]{
\node[circle,draw,fill=n1,inner sep=0pt,minimum size=1.2*\grid] (x1) at ($(#5) + (#1)$) {1};
\node[circle,draw,fill=n2,inner sep=0pt,minimum size=1.2*\grid] (x2) at ($(#5) + (#2)$) {2};
\node[circle,draw,fill=n3,inner sep=0pt,minimum size=1.2*\grid] (x3) at ($(#5) + (#3)$) {3};
\node[circle,draw,fill=n4,inner sep=0pt,minimum size=1.2*\grid] (x4) at ($(#5) + (#4)$) {4};
}

\newcommand{\edges}[8]{
\draw[->,-stealth,draw=e12] (x1) to[bend #1] node[left]{} (x2);
\draw[->,-stealth,draw=e23] (x2) to[bend #2] node[left]{} (x3);
\draw[->,-stealth,draw=e24] (x2) to[bend #3] node[left]{} (x4);
\draw[->,-stealth,draw=e34] (x3) to[bend #4] node[left]{} (x4);
\draw[->,-stealth,draw=e21] (x2) to[bend #5] node[left]{} (x1);
\draw[->,-stealth,draw=e31] (x3) to[bend #6] node[left]{} (x1);
\draw[->,-stealth,draw=e41] (x4) to[bend #7] node[left]{} (x1);
\draw[->,-stealth,draw=e43] (x4) to[bend #8] node[left]{} (x3);
}

\newcommand{\representation}[7]{
\coordinate (p1) at ($(#5) + (-3.0,+8.3)$);
\coordinate (p2) at ($(#5) + (-1.0,+8.3)$);

\draw[step=\grid, black] ($(p1) + (0,0)$) grid ($(p1) + (1,-4)$) node[label={[xshift=-0.5*\grid, yshift=3.3*\grid]\footnotesize #6}] {};
\draw[step=\grid, black] ($(p2) + (0,0)$) grid ($(p2) + (4,-4)$) node[label={[xshift=-2.0*\grid, yshift=3.3*\grid]\footnotesize #7}] {};

\draw[draw=black,fill=n1] ($(p1) + (0,-{#1})$) rectangle ($(p1) + (1,-{#1}) + (0,-1)$);
\draw[draw=black,fill=n2] ($(p1) + (0,-{#2})$) rectangle ($(p1) + (1,-{#2}) + (0,-1)$);
\draw[draw=black,fill=n3] ($(p1) + (0,-{#3})$) rectangle ($(p1) + (1,-{#3}) + (0,-1)$);
\draw[draw=black,fill=n4] ($(p1) + (0,-{#4})$) rectangle ($(p1) + (1,-{#4}) + (0,-1)$);

\draw[draw=black,fill=c00] ($(p2) + ({#1},-{#1})$) rectangle ($(p2) + ({#1},-{#1}) + (1,-1)$);
\draw[draw=black,fill=e21] ($(p2) + ({#1},-{#2})$) rectangle ($(p2) + ({#1},-{#2}) + (1,-1)$);
\draw[draw=black,fill=e31] ($(p2) + ({#1},-{#3})$) rectangle ($(p2) + ({#1},-{#3}) + (1,-1)$);
\draw[draw=black,fill=e41] ($(p2) + ({#1},-{#4})$) rectangle ($(p2) + ({#1},-{#4}) + (1,-1)$);

\draw[draw=black,fill=e12] ($(p2) + ({#2},-{#1})$) rectangle ($(p2) + ({#2},-{#1}) + (1,-1)$);
\draw[draw=black,fill=c00] ($(p2) + ({#2},-{#2})$) rectangle ($(p2) + ({#2},-{#2}) + (1,-1)$);
\draw[draw=black,fill=e32] ($(p2) + ({#2},-{#3})$) rectangle ($(p2) + ({#2},-{#3}) + (1,-1)$);
\draw[draw=black,fill=e42] ($(p2) + ({#2},-{#4})$) rectangle ($(p2) + ({#2},-{#4}) + (1,-1)$);

\draw[draw=black,fill=e13] ($(p2) + ({#3},-{#1})$) rectangle ($(p2) + ({#3},-{#1}) + (1,-1)$);
\draw[draw=black,fill=e23] ($(p2) + ({#3},-{#2})$) rectangle ($(p2) + ({#3},-{#2}) + (1,-1)$);
\draw[draw=black,fill=c00] ($(p2) + ({#3},-{#3})$) rectangle ($(p2) + ({#3},-{#3}) + (1,-1)$);
\draw[draw=black,fill=e43] ($(p2) + ({#3},-{#4})$) rectangle ($(p2) + ({#3},-{#4}) + (1,-1)$);

\draw[draw=black,fill=e14] ($(p2) + ({#4},-{#1})$) rectangle ($(p2) + ({#4},-{#1}) + (1,-1)$);
\draw[draw=black,fill=e24] ($(p2) + ({#4},-{#2})$) rectangle ($(p2) + ({#4},-{#2}) + (1,-1)$);
\draw[draw=black,fill=e34] ($(p2) + ({#4},-{#3})$) rectangle ($(p2) + ({#4},-{#3}) + (1,-1)$);
\draw[draw=black,fill=c00] ($(p2) + ({#4},-{#4})$) rectangle ($(p2) + ({#4},-{#4}) + (1,-1)$);
}

\newcommand{\binarysubgraphselect}[3]{
\node[circle,draw=blue,thick,inner sep=0pt,minimum size=1.2*\grid] (x1) at ($(#3) + (#1)$) {};
\node[circle,draw=blue,thick,inner sep=0pt,minimum size=1.2*\grid] (x2) at ($(#3) + (#2)$) {};
}

\newcommand{\binaryrepresentationselect}[3]{
\coordinate (p1) at ($(#3) + (-3.0,+8.3)$);
\coordinate (p2) at ($(#3) + (-1.0,+8.3)$);

\draw[draw=blue,thick] ($(p1) + (0,-{#1})$) rectangle ($(p1) + (1,-{#1}) + (0,-1)$);
\draw[draw=blue,thick] ($(p1) + (0,-{#2})$) rectangle ($(p1) + (1,-{#2}) + (0,-1)$);

\draw[draw=blue,thick] ($(p2) + ({#1},-{#1})$) rectangle ($(p2) + ({#1},-{#1}) + (1,-1)$);
\draw[draw=blue,thick] ($(p2) + ({#1},-{#2})$) rectangle ($(p2) + ({#1},-{#2}) + (1,-1)$);

\draw[draw=blue,thick] ($(p2) + ({#2},-{#1})$) rectangle ($(p2) + ({#2},-{#1}) + (1,-1)$);
\draw[draw=blue,thick] ($(p2) + ({#2},-{#2})$) rectangle ($(p2) + ({#2},-{#2}) + (1,-1)$);
}

\newcommand{\binarysubgraph}[9]{
\node[circle,draw,fill=#1,inner sep=0pt,minimum size=1.2*\grid] (x1) at ($(#4) + (#5)$) {#8};
\node[circle,draw,fill=#2,inner sep=0pt,minimum size=1.2*\grid] (x2) at ($(#4) + (#6)$) {#9};
\draw[#3] (x1) to[bend #7] node[left]{} (x2);
}

\newcommand{\binaryrepresentation}[7]{
\coordinate (p1) at ($(#5) + (-3.0,+8.3)$);
\coordinate (p2) at ($(#5) + (-1.0,+8.3)$);

\draw[step=\grid, black] ($(p1) + (0,0)$) grid ($(p1) + (1,-2)$) node[label={[xshift=-0.5*\grid, yshift=1.3*\grid]\small #6}] {};
\draw[step=\grid, black] ($(p2) + (0,0)$) grid ($(p2) + (2,-2)$) node[label={[xshift=-1.0*\grid, yshift=1.3*\grid]\small #7}] {};

\draw[draw=black,fill=#1] ($(p1) + (0,-0)$) rectangle ($(p1) + (1,-1)$);
\draw[draw=black,fill=#2] ($(p1) + (0,-1)$) rectangle ($(p1) + (1,-2)$);

\draw[draw=black,fill=c00] ($(p2) + (0,-0)$) rectangle ($(p2) + (1,-1)$);
\draw[draw=black,fill=#3]  ($(p2) + (0,-1)$) rectangle ($(p2) + (1,-2)$);

\draw[draw=black,fill=#4]  ($(p2) + (1,-0)$) rectangle ($(p2) + (2,-1)$);
\draw[draw=black,fill=c00] ($(p2) + (1,-1)$) rectangle ($(p2) + (2,-2)$);
}

\hspace{-80pt}
\begin{tikzpicture}[font=\fontsize{5}{3}\selectfont,x=\grid,y=\grid]

\coordinate (a) at (-20.0,  0);
\coordinate (b) at ( 15.0,  0);
\coordinate (c) at ( 15.0,-18);
\coordinate (d) at (-20.0,-18);

\coordinate (ua) at (-1.9,11.0);
\coordinate (ub) at (+2.0,12.0);
\coordinate (uc) at (+0.0,13.5);
\coordinate (ud) at (-0.5,15.5);

\coordinate (a1) at ($(a) + (-22,0)$);
\coordinate (a2) at ($(a) + (-11,0)$);
\coordinate (a3) at ($(a) + (  0,0)$);
\coordinate (a4) at ($(a) + (  7,0)$);
\coordinate (a5) at ($(a) + ( 14,0)$);

\nodes{ua}{ub}{uc}{ud}{a1}
\edges{left}{right}{right}{right}{right}{right}{right}{left}
\representation{\va}{\vb}{\vc}{\vd}{a1}{$\pi^1\!X$}{$\pi^1\!A$}
\node at ($(a1) + (0.0,3.0)$) {\footnotesize $p^{\text{spn}}_{m,n}(\pi^1\!X,\pi^1\!A)$};

\nodes{ub}{ua}{uc}{ud}{a2}
\edges{left}{left}{left}{right}{right}{left}{left}{left}
\representation{\vb}{\va}{\vc}{\vd}{a2}{$\pi^2\!X$}{$\pi^2\!A$}
\node at ($(a2) + (0.0,3.0)$) {\footnotesize $p^{\text{spn}}_{m,n}(\pi^2 X,\pi^2 A)$};

\nodes{ua}{uc}{ub}{ud}{a3}
\edges{left}{right}{right}{right}{right}{left}{right}{left}
\representation{\va}{\vc}{\vb}{\vd}{a3}{$\pi^3\!X$}{$\pi^3\!A$}
\node at ($(a3) + (0.0,3.0)$) {\footnotesize $p^{\text{spn}}_{m,n}(\pi^3 X,\pi^3 A)$};

\node at ($(a4) + (0.0,6.2)$) {\footnotesize $\cdots$};

\nodes{ua}{uc}{ud}{ub}{a5}
\edges{right}{right}{left}{left}{left}{right}{left}{right}
\representation{\va}{\vc}{\vd}{\vb}{a5}{$\pi^{\textcolor{red}{n!}}\!X$}{$\pi^{\textcolor{red}{n!}}\!A$}
\node at ($(a5) + (0.0,3.0)$) {\footnotesize $p^{\text{spn}}_{m,n}(\pi^{\textcolor{red}{n!}} X,\pi^{\textcolor{red}{n!}} A)$};

\node at ($(a) + (-5.0,0.3)$) {\large (a) Exact invariance};

\coordinate (ba)  at ($(b) + (0,0)$);
\coordinate (ba1) at ($(ba) + (-4,0)$);
\coordinate (ba2) at ($(ba) + (-1,0)$);
\coordinate (ba3) at ($(ba) + ( 5,0)$);

\coordinate (bb)  at ($(b) + (18,0)$);
\coordinate (bb1) at ($(bb) + (-4,0)$);
\coordinate (bb2) at ($(bb) + (-1,0)$);
\coordinate (bb3) at ($(bb) + ( 5,0)$);

\coordinate (bc)  at ($(b) + (28,0)$);

\coordinate (bd)  at ($(b) + (38,0)$);
\coordinate (bd1) at ($(bd) + (-4,0)$);
\coordinate (bd2) at ($(bd) + (-1,0)$);
\coordinate (bd3) at ($(bd) + ( 5,0)$);

\nodes{ud}{uc}{ua}{ub}{ba1}
\edges{left}{right}{left}{right}{right}{left}{right}{left}
\representation{\vd}{\vc}{\va}{\vb}{ba1}{$X$}{$A$}

\binarysubgraphselect{ud}{uc}{ba1}
\binaryrepresentationselect{\vd}{\vc}{ba1}

\binarysubgraph{n1}{n2}{>=stealth,<->,draw=e12}{ba3}{ud}{uc}{left}{1}{2}
\binaryrepresentation{n2}{n1}{e12}{e21}{ba3}{$X_{\mathbf{t}^1}$}{$A_{\mathbf{t}^1}$}
\node at ($(ba3) + (-0.9,5.0)$) {\footnotesize $p^{\text{spn}}_{2,n}(X_{\mathbf{t}^1}, A_{\mathbf{t}^1})$};

\nodes{ud}{uc}{ua}{ub}{bb1}
\edges{left}{right}{left}{right}{right}{left}{right}{left}
\representation{\vd}{\vc}{\va}{\vb}{bb1}{$X$}{$A$}

\binarysubgraphselect{ud}{ua}{bb1}
\binaryrepresentationselect{\vd}{\va}{bb1}

\binarysubgraph{n1}{n3}{>=stealth,<-,draw=e31}{bb3}{ud}{ua}{right}{1}{3}
\binaryrepresentation{n3}{n1}{e13}{e31}{bb3}{$X_{\mathbf{t}^2}$}{$A_{\mathbf{t}^2}$}
\node at ($(bb3) + (-0.9,5.0)$) {\footnotesize $p^{\text{spn}}_{2,n}(X_{\mathbf{t}^2}, A_{\mathbf{t}^2})$};

\node at ($(bc) + (0.0,6.2)$) {\footnotesize $\cdots$};

\nodes{ud}{uc}{ua}{ub}{bd1}
\edges{left}{right}{left}{right}{right}{left}{right}{left}
\representation{\vd}{\vc}{\va}{\vb}{bd1}{$X$}{$A$}

\binarysubgraphselect{uc}{ua}{bd1}
\binaryrepresentationselect{\vc}{\va}{bd1}

\binarysubgraph{n3}{n2}{>=stealth,<-,draw=e23}{bd3}{ua}{uc}{left}{3}{2}
\binaryrepresentation{n3}{n2}{e23}{e32}{bd3}{$X_{\mathbf{t}^{\textcolor{red}{M}}}$}{$A_{\mathbf{t}^{\textcolor{red}{M}}}$}
\node at ($(bd3) + (-0.4,5.0)$) {\footnotesize $p^{\text{spn}}_{2,n}(X_{\mathbf{t}^{\textcolor{red}{M}}}, A_{\mathbf{t}^{\textcolor{red}{M}}})$};

\node at ($(b) + (16.0, 1.3)$) {\large (b) $k$-ary invariance (e.g., binary for $k=2$)};

\coordinate (ca)  at ($(c) + (0,0)$);
\coordinate (ca1) at ($(ca) + (-4,0)$);
\coordinate (ca2) at ($(ca) + (-0,0)$);
\coordinate (ca3) at ($(ca) + ( 4,0)$);

\coordinate (cb)  at ($(c) + (18,0)$);
\coordinate (cb1) at ($(cb) + (-4,0)$);
\coordinate (cb2) at ($(cb) + (-0,0)$);
\coordinate (cb3) at ($(cb) + ( 4,0)$);

\coordinate (cc)  at ($(c) + (28,0)$);

\coordinate (cd)  at ($(c) + (38,0)$);
\coordinate (cd1) at ($(cd) + (-4,0)$);
\coordinate (cd2) at ($(cd) + (-0,0)$);
\coordinate (cd3) at ($(cd) + ( 4,0)$);

\nodes{ud}{uc}{ua}{ub}{ca1}
\edges{left}{right}{left}{right}{right}{left}{right}{left}
\representation{\vd}{\vc}{\va}{\vb}{ca1}{$\pi X$}{$\pi A$}
\node at ($(ca1) + (0.0,3.2)$) {\footnotesize $(\pi X,\pi A)$};

\node at ($(ca2) + (0.0,6.2)$) {\footnotesize\textcolor{red}{$\bm{\rightarrow}$}};

\nodes{ua}{ub}{uc}{ud}{ca3}
\edges{left}{right}{right}{right}{right}{right}{right}{left}
\representation{\va}{\vb}{\vc}{\vd}{ca3}{$\overline{X}$}{$\overline{A}$}
\node at ($(ca3) + (0.0,3.2)$) {\footnotesize $\mathtt{sort}(\pi X,\pi A)$};

\nodes{ub}{ud}{ua}{uc}{cb1}
\edges{right}{right}{left}{right}{left}{right}{left}{left}
\representation{\vb}{\vd}{\va}{\vc}{cb1}{$\pi X$}{$\pi A$}
\node at ($(cb1) + (0.0,3.2)$) {\footnotesize $(\pi X,\pi A)$};

\node at ($(cb2) + (0.0,6.2)$) {\footnotesize\textcolor{red}{$\bm{\rightarrow}$}};

\nodes{ua}{ub}{uc}{ud}{cb3}
\edges{left}{right}{right}{right}{right}{right}{right}{left}
\representation{\va}{\vb}{\vc}{\vd}{cb3}{$\overline{X}$}{$\overline{A}$}
\node at ($(cb3) + (0.0,3.2)$) {\footnotesize $\mathtt{sort}(\pi X,\pi A)$};

\node at ($(cc) + (0.0,6.2)$) {\footnotesize $\cdots$};

\nodes{ud}{ua}{ub}{uc}{cd1}
\edges{right}{right}{left}{left}{left}{right}{left}{left}
\representation{\vd}{\va}{\vb}{\vc}{cd1}{$\pi X$}{$\pi A$}
\node at ($(cd1) + (0.0,3.2)$) {\footnotesize $(\pi X,\pi A)$};

\node at ($(cd2) + (0.0,6.2)$) {\footnotesize\textcolor{red}{$\bm{\rightarrow}$}};

\nodes{ua}{ub}{uc}{ud}{cd3}
\edges{left}{right}{right}{right}{right}{right}{right}{left}
\representation{\va}{\vb}{\vc}{\vd}{cd3}{$\overline{X}$}{$\overline{A}$}
\node at ($(cd3) + (0.0,3.2)$) {\footnotesize $\mathtt{sort}(\pi X,\pi A)$};

\node at ($(c) + (16.0,0.3)$) {\large (d) Sorting};

\coordinate (d1) at ($(d) + (-22,0)$);
\coordinate (d2) at ($(d) + (-11,0)$);
\coordinate (d3) at ($(d) + (  0,0)$);
\coordinate (d4) at ($(d) + (  7,0)$);
\coordinate (d5) at ($(d) + ( 14,0)$);

\nodes{ud}{ub}{uc}{ua}{d1}
\edges{left}{right}{right}{right}{right}{right}{left}{left}
\representation{\vd}{\vb}{\vc}{\va}{d1}{$\pi^1\!X$}{$\pi^1\!A$}
\node at ($(d1) + (0.0,3.0)$) {\footnotesize $p^{\text{spn}}_{m,n}(\pi^1\!X,\pi^1\!A)$};

\nodes{uc}{ud}{ua}{ub}{d2}
\edges{left}{right}{left}{right}{right}{left}{left}{left}
\representation{\vc}{\vd}{\va}{\vb}{d2}{$\pi^2 X$}{$\pi^2 A$}
\node at ($(d2) + (0.0,3.0)$) {\footnotesize $p^{\text{spn}}_{m,n}(\pi^2 X,\pi^2 A)$};

\nodes{uc}{ua}{ud}{ub}{d3}
\edges{left}{left}{right}{left}{right}{left}{right}{right}
\representation{\vc}{\va}{\vd}{\vb}{d3}{$\pi^3 X$}{$\pi^3 A$}
\node at ($(d3) + (0.0,3.0)$) {\footnotesize $p^{\text{spn}}_{m,n}(\pi^3 X,\pi^3 A)$};

\node at ($(d4) + (0.0,6.2)$) {\footnotesize $\cdots$};

\nodes{ud}{ub}{ua}{uc}{d5}
\edges{left}{left}{left}{left}{right}{left}{left}{right}
\representation{\vd}{\vb}{\va}{\vc}{d5}{$\pi^{\textcolor{red}{N}}\!X$}{$\pi^{\textcolor{red}{N}}\!A$}
\node at ($(d5) + (0.0,3.0)$) {\footnotesize $p^{\text{spn}}_{m,n}(\pi^{\textcolor{red}{N}}\!X,\pi^{\textcolor{red}{N}}\!A)$};

\node at ($(d) + (-5.0,0.3)$) {\large (c) Random sampling};

\end{tikzpicture}

%% file: inference.tikz
\newcommand\grid{6.5pt}

\definecolor{c0}{RGB}{243,246,249}
\definecolor{c1}{RGB}{27,158,119}
\definecolor{c2}{RGB}{117,112,179}
\definecolor{c3}{RGB}{217,95,2}
\definecolor{c4}{RGB}{231,41,138}
\definecolor{c5}{RGB}{150,150,150}

\definecolor{n1}{RGB}{252,197,192}
\definecolor{n2}{RGB}{247,104,161}
\definecolor{n3}{RGB}{174,1,126}
\definecolor{n4}{RGB}{122,1,119}

\definecolor{e12}{RGB}{140,81,10}
\definecolor{e13}{RGB}{150,150,150}
\definecolor{e14}{RGB}{150,150,150}
\definecolor{e23}{RGB}{128,205,193}
\definecolor{e24}{RGB}{90,180,172}
\definecolor{e34}{RGB}{1,102,94}

\definecolor{e21}{RGB}{140,81,10}
\definecolor{e31}{RGB}{216,179,101}
\definecolor{e41}{RGB}{246,232,195}
\definecolor{e32}{RGB}{150,150,150}
\definecolor{e42}{RGB}{150,150,150}
\definecolor{e43}{RGB}{1,102,94}

\definecolor{c00}{RGB}{150,150,150}

\newcommand\va{0}
\newcommand\vb{1}
\newcommand\vc{2}
\newcommand\vd{3}

\newcommand{\nodes}[5]{
\node[circle,draw,fill=n1,inner sep=0pt,minimum size=1.2*\grid] (x1) at ($(#5) + (#1)$) {1};
\node[circle,draw,fill=n2,inner sep=0pt,minimum size=1.2*\grid] (x2) at ($(#5) + (#2)$) {2};
\node[circle,draw,fill=n3,inner sep=0pt,minimum size=1.2*\grid] (x3) at ($(#5) + (#3)$) {3};
\node[circle,draw,fill=n4,inner sep=0pt,minimum size=1.2*\grid] (x4) at ($(#5) + (#4)$) {4};
}

\newcommand{\edges}[8]{
\draw[->,-stealth,draw=e12] (x1) to[bend #1] node[left]{} (x2);
\draw[->,-stealth,draw=e23] (x2) to[bend #2] node[left]{} (x3);
\draw[->,-stealth,draw=e24] (x2) to[bend #3] node[left]{} (x4);
\draw[->,-stealth,draw=e34] (x3) to[bend #4] node[left]{} (x4);
\draw[->,-stealth,draw=e21] (x2) to[bend #5] node[left]{} (x1);
\draw[->,-stealth,draw=e31] (x3) to[bend #6] node[left]{} (x1);
\draw[->,-stealth,draw=e41] (x4) to[bend #7] node[left]{} (x1);
\draw[->,-stealth,draw=e43] (x4) to[bend #8] node[left]{} (x3);
}

\newcommand{\representation}[7]{
\coordinate (p1) at ($(#5) + (-3.0,+8.3)$);
\coordinate (p2) at ($(#5) + (-1.0,+8.3)$);

\draw[step=\grid, black] ($(p1) + (0,0)$) grid ($(p1) + (1,-4)$) node[label={[xshift=-0.5*\grid, yshift=3.3*\grid]\footnotesize #6}] {};
\draw[step=\grid, black] ($(p2) + (0,0)$) grid ($(p2) + (4,-4)$) node[label={[xshift=-2.0*\grid, yshift=3.3*\grid]\footnotesize #7}] {};

\draw[draw=black,fill=n1] ($(p1) + (0,-{#1})$) rectangle ($(p1) + (1,-{#1}) + (0,-1)$);
\draw[draw=black,fill=n2] ($(p1) + (0,-{#2})$) rectangle ($(p1) + (1,-{#2}) + (0,-1)$);
\draw[draw=black,fill=n3] ($(p1) + (0,-{#3})$) rectangle ($(p1) + (1,-{#3}) + (0,-1)$);
\draw[draw=black,fill=n4] ($(p1) + (0,-{#4})$) rectangle ($(p1) + (1,-{#4}) + (0,-1)$);

\draw[draw=black,fill=c00] ($(p2) + ({#1},-{#1})$) rectangle ($(p2) + ({#1},-{#1}) + (1,-1)$);
\draw[draw=black,fill=e21] ($(p2) + ({#1},-{#2})$) rectangle ($(p2) + ({#1},-{#2}) + (1,-1)$);
\draw[draw=black,fill=e31] ($(p2) + ({#1},-{#3})$) rectangle ($(p2) + ({#1},-{#3}) + (1,-1)$);
\draw[draw=black,fill=e41] ($(p2) + ({#1},-{#4})$) rectangle ($(p2) + ({#1},-{#4}) + (1,-1)$);

\draw[draw=black,fill=e12] ($(p2) + ({#2},-{#1})$) rectangle ($(p2) + ({#2},-{#1}) + (1,-1)$);
\draw[draw=black,fill=c00] ($(p2) + ({#2},-{#2})$) rectangle ($(p2) + ({#2},-{#2}) + (1,-1)$);
\draw[draw=black,fill=e32] ($(p2) + ({#2},-{#3})$) rectangle ($(p2) + ({#2},-{#3}) + (1,-1)$);
\draw[draw=black,fill=e42] ($(p2) + ({#2},-{#4})$) rectangle ($(p2) + ({#2},-{#4}) + (1,-1)$);

\draw[draw=black,fill=e13] ($(p2) + ({#3},-{#1})$) rectangle ($(p2) + ({#3},-{#1}) + (1,-1)$);
\draw[draw=black,fill=e23] ($(p2) + ({#3},-{#2})$) rectangle ($(p2) + ({#3},-{#2}) + (1,-1)$);
\draw[draw=black,fill=c00] ($(p2) + ({#3},-{#3})$) rectangle ($(p2) + ({#3},-{#3}) + (1,-1)$);
\draw[draw=black,fill=e43] ($(p2) + ({#3},-{#4})$) rectangle ($(p2) + ({#3},-{#4}) + (1,-1)$);

\draw[draw=black,fill=e14] ($(p2) + ({#4},-{#1})$) rectangle ($(p2) + ({#4},-{#1}) + (1,-1)$);
\draw[draw=black,fill=e24] ($(p2) + ({#4},-{#2})$) rectangle ($(p2) + ({#4},-{#2}) + (1,-1)$);
\draw[draw=black,fill=e34] ($(p2) + ({#4},-{#3})$) rectangle ($(p2) + ({#4},-{#3}) + (1,-1)$);
\draw[draw=black,fill=c00] ($(p2) + ({#4},-{#4})$) rectangle ($(p2) + ({#4},-{#4}) + (1,-1)$);
}

\newcommand{\subgraphtargetone}[6]{
\node[circle,draw,fill=blue,inner sep=0pt,minimum size=1.2*\grid] (x2) at ($(#1) + (#2)$) {2};

\draw[->,-stealth,draw=blue] (x1) to[bend #3] node[left]{} (x2);
\draw[->,-stealth,draw=blue] (x2) to[bend #4] node[left]{} (x3);
\draw[->,-stealth,draw=blue] (x2) to[bend #5] node[left]{} (x4);
\draw[->,-stealth,draw=blue] (x2) to[bend #6] node[left]{} (x1);
}

\newcommand{\representationtargetone}[5]{
\coordinate (p1) at ($(#5) + (-3.0,+8.3)$);
\coordinate (p2) at ($(#5) + (-1.0,+8.3)$);

\draw[draw=black,fill=blue] ($(p1) + (0,-{#2})$) rectangle ($(p1) + (1,-{#2}) + (0,-1)$);

\draw[draw=black,fill=blue] ($(p2) + ({#1},-{#2})$) rectangle ($(p2) + ({#1},-{#2}) + (1,-1)$);

\draw[draw=black,fill=blue] ($(p2) + ({#2},-{#1})$) rectangle ($(p2) + ({#2},-{#1}) + (1,-1)$);
\draw[draw=black,fill=blue] ($(p2) + ({#2},-{#2})$) rectangle ($(p2) + ({#2},-{#2}) + (1,-1)$);
\draw[draw=black,fill=blue] ($(p2) + ({#2},-{#3})$) rectangle ($(p2) + ({#2},-{#3}) + (1,-1)$);
\draw[draw=black,fill=blue] ($(p2) + ({#2},-{#4})$) rectangle ($(p2) + ({#2},-{#4}) + (1,-1)$);

\draw[draw=black,fill=blue] ($(p2) + ({#3},-{#2})$) rectangle ($(p2) + ({#3},-{#2}) + (1,-1)$);

\draw[draw=black,fill=blue] ($(p2) + ({#4},-{#2})$) rectangle ($(p2) + ({#4},-{#2}) + (1,-1)$);
}

\newcommand{\nodesmarginal}[5]{
\node[circle,draw,fill=n1,inner sep=0pt,minimum size=1.2*\grid] (x1) at ($(#5) + (#1)$) {1};
\node[circle,draw,fill=n3,inner sep=0pt,minimum size=1.2*\grid] (x3) at ($(#5) + (#3)$) {3};
\node[circle,draw,fill=n4,inner sep=0pt,minimum size=1.2*\grid] (x4) at ($(#5) + (#4)$) {4};
}

\newcommand{\edgesmarginal}[8]{
\draw[->,-stealth,draw=e34] (x3) to[bend #4] node[left]{} (x4);
\draw[->,-stealth,draw=e31] (x3) to[bend #6] node[left]{} (x1);
\draw[->,-stealth,draw=e41] (x4) to[bend #7] node[left]{} (x1);
\draw[->,-stealth,draw=e43] (x4) to[bend #8] node[left]{} (x3);
}

\newcommand{\representationmarginal}[7]{
\coordinate (p1) at ($(#5) + (-3.0,+8.3)$);
\coordinate (p2) at ($(#5) + (-1.0,+8.3)$);

\draw[step=\grid, black] ($(p1) + (0,0)$) grid ($(p1) + (1,-3)$) node[label={[xshift=-0.5*\grid, yshift=2.1*\grid]\footnotesize #6}] {};
\draw[step=\grid, black] ($(p2) + (0,0)$) grid ($(p2) + (3,-3)$) node[label={[xshift=-1.5*\grid, yshift=2.1*\grid]\footnotesize #7}] {};

\draw[draw=black,fill=n1] ($(p1) + (0,-{#1})$) rectangle ($(p1) + (1,-{#1}) + (0,-1)$);
\draw[draw=black,fill=n3] ($(p1) + (0,-{#2})$) rectangle ($(p1) + (1,-{#2}) + (0,-1)$);
\draw[draw=black,fill=n4] ($(p1) + (0,-{#3})$) rectangle ($(p1) + (1,-{#3}) + (0,-1)$);

\draw[draw=black,fill=c00] ($(p2) + ({#1},-{#1})$) rectangle ($(p2) + ({#1},-{#1}) + (1,-1)$);
\draw[draw=black,fill=e31] ($(p2) + ({#1},-{#2})$) rectangle ($(p2) + ({#1},-{#2}) + (1,-1)$);
\draw[draw=black,fill=e41] ($(p2) + ({#1},-{#3})$) rectangle ($(p2) + ({#1},-{#3}) + (1,-1)$);

\draw[draw=black,fill=e13] ($(p2) + ({#2},-{#1})$) rectangle ($(p2) + ({#2},-{#1}) + (1,-1)$);
\draw[draw=black,fill=c00] ($(p2) + ({#2},-{#2})$) rectangle ($(p2) + ({#2},-{#2}) + (1,-1)$);
\draw[draw=black,fill=e43] ($(p2) + ({#2},-{#3})$) rectangle ($(p2) + ({#2},-{#3}) + (1,-1)$);

\draw[draw=black,fill=e14] ($(p2) + ({#3},-{#1})$) rectangle ($(p2) + ({#3},-{#1}) + (1,-1)$);
\draw[draw=black,fill=e34] ($(p2) + ({#3},-{#2})$) rectangle ($(p2) + ({#3},-{#2}) + (1,-1)$);
\draw[draw=black,fill=c00] ($(p2) + ({#3},-{#3})$) rectangle ($(p2) + ({#3},-{#3}) + (1,-1)$);
}

\hspace{-80pt}
\begin{tikzpicture}[font=\fontsize{5}{3}\selectfont,x=\grid,y=\grid]
\coordinate (ua) at (-1.9,11.0);
\coordinate (ub) at (+2.0,12.0);
\coordinate (uc) at (+0.0,13.5);
\coordinate (ud) at (-0.5,15.5);

\coordinate (a)  at (-55,9);
\coordinate (b)  at (-15,0);
\coordinate (c)  at ( 18,0);

\node at (a) {\footnotesize
$\begin{aligned}
f(X,A)
&=
\textcolor{blue}{h(X_2)h(A_{22})\prod_{k\notin\mathbf{a}}h(A_{2k})}
\prod_{t\notin\mathbf{a}}h(X_t)
    \bigg(
    \prod_{u\notin\mathbf{a}}h(A_{tu})
    \textcolor{blue}{h(A_{t2})}
    \bigg)
\\
&=
\textcolor{blue}{\mathds{1}_\mathcal{X}(X_2)\mathds{1}_\mathcal{A}(A_{22})\prod_{k\notin\mathbf{a}}\mathds{1}_\mathcal{A}(A_{2k})}
\prod_{t\notin\mathbf{a}}\mathds{1}_{\overline{X}_t}(X_t)
    \bigg(
    \prod_{u\notin\mathbf{a}}\mathds{1}_{\overline{A}_{tu}}(A_{tu})
    \textcolor{blue}{\mathds{1}_\mathcal{A}(A_{t2})}
    \bigg)
\end{aligned}$
};

\node at ($(a) + (2.0,-6.5)$) {\large (a) An example of \eqref{eq:f} for $\mathbf{a}\coloneqq\{2\}$};

\nodes{ua}{ub}{uc}{ud}{b}
\edges{left}{right}{right}{right}{right}{right}{right}{left}
\representation{\va}{\vb}{\vc}{\vd}{b}{$X$}{$A$}

\node at ($(b) + (1.0,2.5)$) {\large (b) The corresponding node of $G$};

\representationtargetone{\va}{\vb}{\vc}{\vd}{b}
\subgraphtargetone{b}{ub}{left}{right}{right}{right}

\nodesmarginal{ua}{ub}{uc}{ud}{c}
\edgesmarginal{left}{right}{right}{right}{right}{right}{right}{left}
\representationmarginal{\va}{\vb}{\vc}{\vd}{c}{$\overline{X}$}{$\overline{A}$}

\node at ($(c) + (-2.0,2.5)$) {\large (c) $G$ after the marginal query};

\end{tikzpicture}

%% file: tpm2024.bbl
\begin{thebibliography}{45}
\providecommand{\natexlab}[1]{#1}
\providecommand{\url}[1]{\texttt{#1}}
\expandafter\ifx\csname urlstyle\endcsname\relax
  \providecommand{\doi}[1]{doi: #1}\else
  \providecommand{\doi}{doi: \begingroup \urlstyle{rm}\Url}\fi

\bibitem[Barndorff-Nielsen(1978)]{barndorff1978information}
Ole Barndorff-Nielsen.
\newblock \emph{Information and exponential families: In statistical theory}.
\newblock John Wiley \& Sons, 1978.

\bibitem[Chandak et~al.(2023)Chandak, Huang, and Zitnik]{chandak2023building}
Payal Chandak, Kexin Huang, and Marinka Zitnik.
\newblock Building a knowledge graph to enable precision medicine.
\newblock \emph{Scientific Data}, 10\penalty0 (1):\penalty0 67, 2023.

\bibitem[Choi et~al.(2020)Choi, Vergari, and Van~den Broeck]{choi2020probabilistic}
Y~Choi, Antonio Vergari, and Guy Van~den Broeck.
\newblock Probabilistic circuits: A unifying framework for tractable probabilistic models.
\newblock \emph{UCLA. URL: http://starai. cs. ucla. edu/papers/ProbCirc20. pdf}, 2020.

\bibitem[Choudhary et~al.(2022)Choudhary, DeCost, Chen, Jain, Tavazza, Cohn, Park, Choudhary, Agrawal, Billinge, et~al.]{choudhary2022recent}
Kamal Choudhary, Brian DeCost, Chi Chen, Anubhav Jain, Francesca Tavazza, Ryan Cohn, Cheol~Woo Park, Alok Choudhary, Ankit Agrawal, Simon~JL Billinge, et~al.
\newblock Recent advances and applications of deep learning methods in materials science.
\newblock \emph{npj Computational Materials}, 8\penalty0 (1):\penalty0 59, 2022.

\bibitem[Daley and Vere-Jones(2003)]{daley2003introduction}
Daryl~J Daley and David Vere-Jones.
\newblock \emph{An introduction to the theory of point processes: volume I: elementary theory and methods}.
\newblock Springer, 2003.

\bibitem[De~Cao and Kipf(2018)]{de2018molgan}
Nicola De~Cao and Thomas Kipf.
\newblock Mol{GAN}: An implicit generative model for small molecular graphs.
\newblock \emph{arXiv preprint arXiv:1805.11973}, 2018.

\bibitem[Defferrard et~al.(2016)Defferrard, Bresson, and Vandergheynst]{defferrard2016convolutional}
Micha{\"e}l Defferrard, Xavier Bresson, and Pierre Vandergheynst.
\newblock Convolutional neural networks on graphs with fast localized spectral filtering.
\newblock \emph{Advances in neural information processing systems}, 29, 2016.

\bibitem[Derrow-Pinion et~al.(2021)Derrow-Pinion, She, Wong, Lange, Hester, Perez, Nunkesser, Lee, Guo, Wiltshire, et~al.]{derrow2021eta}
Austin Derrow-Pinion, Jennifer She, David Wong, Oliver Lange, Todd Hester, Luis Perez, Marc Nunkesser, Seongjae Lee, Xueying Guo, Brett Wiltshire, et~al.
\newblock {ETA} prediction with graph neural networks in {G}oogle maps.
\newblock In \emph{Proceedings of the 30th ACM International Conference on Information \& Knowledge Management}, pages 3767--3776, 2021.

\bibitem[Errica and Niepert(2023)]{errica2023tractable}
Federico Errica and Mathias Niepert.
\newblock Tractable probabilistic graph representation learning with graph-induced sum-product networks.
\newblock \emph{arXiv preprint arXiv:2305.10544}, 2023.

\bibitem[G{\'o}mez-Bombarelli et~al.(2018)G{\'o}mez-Bombarelli, Wei, Duvenaud, Hern{\'a}ndez-Lobato, S{\'a}nchez-Lengeling, Sheberla, Aguilera-Iparraguirre, Hirzel, Adams, and Aspuru-Guzik]{gomez2018automatic}
Rafael G{\'o}mez-Bombarelli, Jennifer~N Wei, David Duvenaud, Jos{\'e}~Miguel Hern{\'a}ndez-Lobato, Benjam{\'\i}n S{\'a}nchez-Lengeling, Dennis Sheberla, Jorge Aguilera-Iparraguirre, Timothy~D Hirzel, Ryan~P Adams, and Al{\'a}n Aspuru-Guzik.
\newblock Automatic chemical design using a data-driven continuous representation of molecules.
\newblock \emph{ACS central science}, 4\penalty0 (2):\penalty0 268--276, 2018.

\bibitem[Honda et~al.(2019)Honda, Akita, Ishiguro, Nakanishi, and Oono]{honda2019graph}
Shion Honda, Hirotaka Akita, Katsuhiko Ishiguro, Toshiki Nakanishi, and Kenta Oono.
\newblock Graph residual flow for molecular graph generation.
\newblock \emph{arXiv preprint arXiv:1909.13521}, 2019.

\bibitem[Jin et~al.(2018)Jin, Barzilay, and Jaakkola]{jin2018junction}
Wengong Jin, Regina Barzilay, and Tommi Jaakkola.
\newblock Junction tree variational autoencoder for molecular graph generation.
\newblock In \emph{International conference on machine learning}, pages 2323--2332. PMLR, 2018.

\bibitem[Jo et~al.(2022)Jo, Lee, and Hwang]{jo2022score}
Jaehyeong Jo, Seul Lee, and Sung~Ju Hwang.
\newblock Score-based generative modeling of graphs via the system of stochastic differential equations.
\newblock In \emph{International Conference on Machine Learning}, pages 10362--10383. PMLR, 2022.

\bibitem[Kingma and Ba(2014)]{kingma2014adam}
Diederik~P Kingma and Jimmy Ba.
\newblock Adam: A method for stochastic optimization.
\newblock \emph{arXiv preprint arXiv:1412.6980}, 2014.

\bibitem[Kusner et~al.(2017)Kusner, Paige, and Hern{\'a}ndez-Lobato]{kusner2017grammar}
Matt~J Kusner, Brooks Paige, and Jos{\'e}~Miguel Hern{\'a}ndez-Lobato.
\newblock Grammar variational autoencoder.
\newblock In \emph{International conference on machine learning}, pages 1945--1954. PMLR, 2017.

\bibitem[Landrum et~al.(2006)]{landrum2006rdkit}
Greg Landrum et~al.
\newblock Rdkit: Open-source cheminformatics, 2006.

\bibitem[Loconte et~al.(2024)Loconte, Sladek, Mengel, Trapp, Solin, Gillis, and Vergari]{loconte2024subtractive}
Lorenzo Loconte, Aleksanteri~M Sladek, Stefan Mengel, Martin Trapp, Arno Solin, Nicolas Gillis, and Antonio Vergari.
\newblock Subtractive mixture models via squaring: Representation and learning.
\newblock In \emph{The 12th International Conference on Learning Representations}, 2024.

\bibitem[Luo et~al.(2021)Luo, Yan, and Ji]{luo2021graphdf}
Youzhi Luo, Keqiang Yan, and Shuiwang Ji.
\newblock Graph{DF}: A discrete flow model for molecular graph generation.
\newblock In \emph{International conference on machine learning}, pages 7192--7203. PMLR, 2021.

\bibitem[Ma et~al.(2018)Ma, Chen, and Xiao]{ma2018constrained}
Tengfei Ma, Jie Chen, and Cao Xiao.
\newblock Constrained generation of semantically valid graphs via regularizing variational autoencoders.
\newblock \emph{Advances in Neural Information Processing Systems}, 31, 2018.

\bibitem[Madhawa et~al.(2019)Madhawa, Ishiguro, Nakago, and Abe]{madhawa2019graphnvp}
Kaushalya Madhawa, Katushiko Ishiguro, Kosuke Nakago, and Motoki Abe.
\newblock Graphnvp: An invertible flow model for generating molecular graphs.
\newblock \emph{arXiv preprint arXiv:1905.11600}, 2019.

\bibitem[Montavon et~al.(2012)Montavon, Hansen, Fazli, Rupp, Biegler, Ziehe, Tkatchenko, Lilienfeld, and M{\"u}ller]{montavon2012learning}
Gr{\'e}goire Montavon, Katja Hansen, Siamac Fazli, Matthias Rupp, Franziska Biegler, Andreas Ziehe, Alexandre Tkatchenko, Anatole Lilienfeld, and Klaus-Robert M{\"u}ller.
\newblock Learning invariant representations of molecules for atomization energy prediction.
\newblock \emph{Advances in neural information processing systems}, 25, 2012.

\bibitem[Murphy et~al.(2019{\natexlab{a}})Murphy, Srinivasan, Rao, and Ribeiro]{murphy2019relational}
Ryan Murphy, Balasubramaniam Srinivasan, Vinayak Rao, and Bruno Ribeiro.
\newblock Relational pooling for graph representations.
\newblock In \emph{International Conference on Machine Learning}, pages 4663--4673. PMLR, 2019{\natexlab{a}}.

\bibitem[Murphy et~al.(2019{\natexlab{b}})Murphy, Srinivasan, Rao, and Ribeiro]{murphy2018janossy}
Ryan~L Murphy, Balasubramaniam Srinivasan, Vinayak Rao, and Bruno Ribeiro.
\newblock Janossy pooling: Learning deep permutation-invariant functions for variable-size inputs.
\newblock In \emph{The 7th International Conference on Learning Representations}, 2019{\natexlab{b}}.

\bibitem[Niepert et~al.(2016)Niepert, Ahmed, and Kutzkov]{niepert2016learning}
Mathias Niepert, Mohamed Ahmed, and Konstantin Kutzkov.
\newblock Learning convolutional neural networks for graphs.
\newblock In \emph{International conference on machine learning}, pages 2014--2023. PMLR, 2016.

\bibitem[Orbanz and Roy(2014)]{orbanz2014bayesian}
Peter Orbanz and Daniel~M Roy.
\newblock Bayesian models of graphs, arrays and other exchangeable random structures.
\newblock \emph{IEEE transactions on pattern analysis and machine intelligence}, 37\penalty0 (2):\penalty0 437--461, 2014.

\bibitem[Pape\v{z} et~al.(2024)Pape\v{z}, Rektoris, \v{S}m\'{i}dl, and Pevn{\'y}]{papez2024sum}
Milan Pape\v{z}, Martin Rektoris, V\'{a}clav \v{S}m\'{i}dl, and Tom{\'a}{\v{s}} Pevn{\'y}.
\newblock Sum-product-set networks: Deep tractable models for tree-structured graphs.
\newblock In \emph{The 12th International Conference on Learning Representations}, 2024.

\bibitem[Peharz et~al.(2020{\natexlab{a}})Peharz, Lang, Vergari, Stelzner, Molina, Trapp, Van~den Broeck, Kersting, and Ghahramani]{peharz2020einsum}
Robert Peharz, Steven Lang, Antonio Vergari, Karl Stelzner, Alejandro Molina, Martin Trapp, Guy Van~den Broeck, Kristian Kersting, and Zoubin Ghahramani.
\newblock Einsum networks: {F}ast and scalable learning of tractable probabilistic circuits.
\newblock In \emph{International Conference on Machine Learning}, pages 7563--7574. PMLR, 2020{\natexlab{a}}.

\bibitem[Peharz et~al.(2020{\natexlab{b}})Peharz, Vergari, Stelzner, Molina, Shao, Trapp, Kersting, and Ghahramani]{peharz2020random}
Robert Peharz, Antonio Vergari, Karl Stelzner, Alejandro Molina, Xiaoting Shao, Martin Trapp, Kristian Kersting, and Zoubin Ghahramani.
\newblock Random sum-product networks: A simple and effective approach to probabilistic deep learning.
\newblock In \emph{Uncertainty in Artificial Intelligence}, pages 334--344. PMLR, 2020{\natexlab{b}}.

\bibitem[Poon and Domingos(2011)]{poon2011sum}
Hoifung Poon and Pedro Domingos.
\newblock Sum-product networks: A new deep architecture.
\newblock In \emph{2011 IEEE International Conference on Computer Vision Workshops (ICCV Workshops)}, pages 689--690. IEEE, 2011.

\bibitem[Ramakrishnan et~al.(2014)Ramakrishnan, Dral, Rupp, and Von~Lilienfeld]{ramakrishnan2014quantum}
Raghunathan Ramakrishnan, Pavlo~O Dral, Matthias Rupp, and O~Anatole Von~Lilienfeld.
\newblock Quantum chemistry structures and properties of 134 kilo molecules.
\newblock \emph{Scientific data}, 1\penalty0 (1):\penalty0 1--7, 2014.

\bibitem[Reymond et~al.(2012)Reymond, Ruddigkeit, Blum, and Van~Deursen]{reymond2012enumeration}
Jean-Louis Reymond, Lars Ruddigkeit, Lorenz Blum, and Ruud Van~Deursen.
\newblock The enumeration of chemical space.
\newblock \emph{Wiley Interdisciplinary Reviews: Computational Molecular Science}, 2\penalty0 (5):\penalty0 717--733, 2012.

\bibitem[Shi et~al.(2020)Shi, Xu, Zhu, Zhang, Zhang, and Tang]{shi2020graphaf}
Chence Shi, Minkai Xu, Zhaocheng Zhu, Weinan Zhang, Ming Zhang, and Jian Tang.
\newblock Graph{AF}: A flow-based autoregressive model for molecular graph generation.
\newblock \emph{arXiv preprint arXiv:2001.09382}, 2020.

\bibitem[Shih et~al.(2019)Shih, Van~den Broeck, Beame, and Amarilli]{shih2019smoothing}
Andy Shih, Guy Van~den Broeck, Paul Beame, and Antoine Amarilli.
\newblock Smoothing structured decomposable circuits.
\newblock \emph{Advances in Neural Information Processing Systems}, 32, 2019.

\bibitem[Simonovsky and Komodakis(2018)]{simonovsky2018graphvae}
Martin Simonovsky and Nikos Komodakis.
\newblock Graph{VAE}: {T}owards generation of small graphs using variational autoencoders.
\newblock In \emph{Artificial Neural Networks and Machine Learning--ICANN 2018: 27th International Conference on Artificial Neural Networks, Rhodes, Greece, October 4-7, 2018, Proceedings, Part I 27}, pages 412--422. Springer, 2018.

\bibitem[Veitch and Roy(2015)]{veitch2015class}
Victor Veitch and Daniel~M Roy.
\newblock The class of random graphs arising from exchangeable random measures.
\newblock \emph{arXiv preprint arXiv:1512.03099}, 2015.

\bibitem[Vergari et~al.(2019)Vergari, Di~Mauro, and Esposito]{vergari2019visualizing}
Antonio Vergari, Nicola Di~Mauro, and Floriana Esposito.
\newblock Visualizing and understanding sum-product networks.
\newblock \emph{Machine Learning}, 108\penalty0 (4):\penalty0 551--573, 2019.

\bibitem[Vergari et~al.(2021)Vergari, Choi, Liu, Teso, and Broeck]{vergari2021compositional}
Antonio Vergari, YooJung Choi, Anji Liu, Stefano Teso, and Guy Van~den Broeck.
\newblock A compositional atlas of tractable circuit operations: From simple transformations to complex information-theoretic queries.
\newblock \emph{arXiv preprint arXiv:2102.06137}, 2021.

\bibitem[Verma et~al.(2022)Verma, Kaski, Heinonen, and Garg]{verma2022modular}
Yogesh Verma, Samuel Kaski, Markus Heinonen, and Vikas Garg.
\newblock Modular flows: Differential molecular generation.
\newblock \emph{Advances in neural information processing systems}, 35:\penalty0 12409--12421, 2022.

\bibitem[Wagstaff et~al.(2022)Wagstaff, Fuchs, Engelcke, Osborne, and Posner]{wagstaff2022universal}
Edward Wagstaff, Fabian~B Fuchs, Martin Engelcke, Michael~A Osborne, and Ingmar Posner.
\newblock Universal approximation of functions on sets.
\newblock \emph{Journal of Machine Learning Research}, 23\penalty0 (151):\penalty0 1--56, 2022.

\bibitem[Wu et~al.(2020)Wu, Pan, Chen, Long, Zhang, and Philip]{wu2020comprehensive}
Zonghan Wu, Shirui Pan, Fengwen Chen, Guodong Long, Chengqi Zhang, and S~Yu Philip.
\newblock A comprehensive survey on graph neural networks.
\newblock \emph{IEEE transactions on neural networks and learning systems}, 32\penalty0 (1):\penalty0 4--24, 2020.

\bibitem[You et~al.(2018)You, Ying, Ren, Hamilton, and Leskovec]{you2018graphrnn}
Jiaxuan You, Rex Ying, Xiang Ren, William Hamilton, and Jure Leskovec.
\newblock Graph{RNN}: {G}enerating realistic graphs with deep auto-regressive models.
\newblock In \emph{International conference on machine learning}, pages 5708--5717. PMLR, 2018.

\bibitem[Zaheer et~al.(2017)Zaheer, Kottur, Ravanbakhsh, Poczos, Salakhutdinov, and Smola]{zaheer2017deep}
Manzil Zaheer, Satwik Kottur, Siamak Ravanbakhsh, Barnabas Poczos, Russ~R Salakhutdinov, and Alexander~J Smola.
\newblock Deep sets.
\newblock \emph{Advances in neural information processing systems}, 30, 2017.

\bibitem[Zang and Wang(2020)]{zang2020moflow}
Chengxi Zang and Fei Wang.
\newblock Moflow: An invertible flow model for generating molecular graphs.
\newblock In \emph{Proceedings of the 26th ACM SIGKDD international conference on knowledge discovery \& data mining}, pages 617--626, 2020.

\bibitem[Zhang et~al.(2020)Zhang, Cui, and Zhu]{zhang2020deep}
Ziwei Zhang, Peng Cui, and Wenwu Zhu.
\newblock Deep learning on graphs: A survey.
\newblock \emph{IEEE Transactions on Knowledge and Data Engineering}, 34\penalty0 (1):\penalty0 249--270, 2020.

\bibitem[Zheng et~al.(2018)Zheng, Pronobis, and Rao]{zheng2018learning}
Kaiyu Zheng, Andrzej Pronobis, and Rajesh Rao.
\newblock Learning graph-structured sum-product networks for probabilistic semantic maps.
\newblock In \emph{Proceedings of the AAAI Conference on Artificial Intelligence}, volume~32, 2018.

\end{thebibliography}
